\documentclass[12pt]{spie}  
\usepackage{booktabs}
\usepackage{cite}
\usepackage{makecell}
\usepackage{booktabs}
\usepackage{xcolor}
\usepackage{setspace}
\usepackage{graphicx}
\usepackage{float}
\usepackage{bm}
\usepackage{subcaption}
\usepackage{amsmath}
\usepackage{amssymb}
\usepackage{adjustbox}
\usepackage{multirow}
\usepackage[customcolors]{hf-tikz}
\usepackage[hidelinks]{hyperref}
\usepackage{comment}
\usepackage[title]{appendix}
\hypersetup{
	colorlinks   = true, 
	urlcolor     = blue, 
	linkcolor    = blue, 
	citecolor    = red   
}
\title{Physics-guided Deep Markov Models for Learning Nonlinear Dynamical Systems with Uncertainty}

\author{Wei Liu$^{1,3}$, Zhilu Lai$^{2,3*}$, Kiran Bacsa$^{2,3}$, and Eleni Chatzi$^{2,3}$	\skiplinehalf
	\small $^{1}$
    Department of Industrial Systems Engineering and Management, National University of Singapore, Singapore\\
	\small $^{2}$
	Department of Civil, Environmental and Geomatic Engineering, ETH Z\"urich, Z\"urich, Switzerland\\
    \small $^{3}$
    Future Resilient Systems, Singapore-ETH Centre, Singapore\\
}

\authorinfo{$^{*}$Further author information: send correspondence to laiz@ethz.ch \\
\textit{This manuscript has been accepted for publication in Mechanical Systems and Signal Processing.} }
\begin{document} 
	\maketitle 
	\begin{abstract}

In this paper, we propose a probabilistic physics-guided framework, termed Physics-guided Deep Markov Model (PgDMM). The framework targets the inference of the characteristics and latent structure of nonlinear dynamical systems from measurement data, where exact inference of latent variables is typically intractable. A recently surfaced option pertains to leveraging variational inference to perform approximate inference. In such a scheme, transition and emission functions of the system are parameterized via feed-forward neural networks (deep generative models). However, due to the generalized and highly versatile formulation of neural network functions, the learned latent space often lacks physical interpretation and structured representation. To address this, we bridge physics-based state space models with Deep Markov Models, thus delivering a hybrid modeling framework for unsupervised learning and identification of nonlinear dynamical systems. The proposed framework takes advantage of the expressive power of deep learning, while retaining the driving physics of the dynamical system by imposing physics-driven restrictions on the side of the latent space. We demonstrate the benefits of such a fusion in terms of achieving improved performance on illustrative simulation examples and experimental case studies of nonlinear systems. Our results indicate that the physics-based models involved in the employed transition and emission functions essentially enforce a more structured and physically interpretable latent space, which is essential for enhancing and generalizing the predictive capabilities of deep learning-based models.
	\end{abstract} 
\section{Introduction}
Over the last few decades, machine learning and deep learning techniques \cite{lecun2015deep,goodfellow2016deep}, have been successfully employed in a broad range of challenging applications, such as image/speech recognition \cite{he2016deep,svensen2018deep}, natural language processing \cite{wu2016google,hieber2017sockeye}, and complex systems modeling\cite{raissi2019physics}. Deep learning seeks to deliver a data-driven representation of a system via use of a computational model that is composed of multiple processing layers. Within the context of modeling for dynamical systems in particular, this approach shares some of the bottlenecks and limitations of the classical approach of system identification \cite{ljung2020deep}, where the aim is to \textit{learn or infer a predictive model from measurement data}. Traditional modeling methods typically rely on differential equations and impose assumptions on the inherent structure of the dynamics, leaving only a few parameters as unknown quantities that are to be inferred. On the other hand, the more recently surfaced deep learning schemes \cite{farrar2012structural} are highly expressive and flexible in learning from raw data, and have found broad acceptance in structural dynamics applications \cite{zhang2020machine,xu2021phymdan}. A number of successful applications, which integrate both fields, i.e., structural system identification and machine learning, are found in existing literature \cite{farrar2012structural,dervilis2013machine,jiang2017fuzzy,lai2019sparse,chen2020sparse,liu2021knowledge,lai2021structural,bull2021probabilistic}. A key challenge in this respect lies in learning a model that is capable of extrapolation and generalization, i.e., one that can accurately predict the system's behavior for ranges of inputs that extend beyond the training dataset. This implies that the learned model adequately captures the inherent nature of the system. Particularly in the case of nonlinear systems, the behavior of a system can vary significantly for different loading or initial condition regimes. 

The observations generated from dynamical systems comprise sequential data (time series signals or image sequences), which form the input for system identification tasks. In deep learning literature, such observations are mainly treated via use of recurrent neural network (RNN) architectures and its gated variants (e.g., gated recurrent units, GRU \cite{chung2014empirical}), which offer powerful tools for modeling temporal data. While RNNs form deterministic models, Dynamic Bayesian Networks \cite{koller2009probabilistic,li2017dynamic}, which account for temporal dependencies (such as Hidden Markov Models\cite{rabiner1986introduction}), offer a probabilistic approach for learning the structure of generative models and enjoy widespread adoption for sequential data. Very recently, \textit{Dynamical Variational Autoencoders} \cite{girin2020dynamical} emerged as a sequential version of variational autoencoders (VAE \cite{kingma2013auto}), or as a variational version of Dynamical Bayesian Networks, and have been applied to learning a latent space representation for sequential data in an unsupervised manner {\color{black} to discover the latent quantities governing the dynamical system and capture the dynamics underlying the data}. {\color{black} The VAE models are generally considered as unsupervised, in the sense that they are trained to learn a latent representation from raw data by maximizing the observed data likelihood, while requiring no specification of additional labels or targets for latent features. We additionally wish to clarify that the method we here propose is an input/output learning, or identification method, in the sense that it assumes knowledge of the input (loading) signal. These two concepts, i.e. ``unsupervised learning" and ``input/output identification", are separate.} The VAE models typically parameterize the involved distributions by deep neural networks, which allows for learning high-dimensional and highly multi-modal distributions. {\color{black} Particularly, dynamical VAEs} comprise an inference model (encoder) and a generative model (decoder) and adopt a variational or evidence lower bound (ELBO) maximization methodology, for deriving a best fit model of the dynamical system. Several variants of this class have been proposed, by assuming different schemes to represent the interdependence between the latent and the observed variables, namely Deep Markov Models (DMM) \cite{krishnan2015deep,krishnan2017structured}, Deep Variational Bayes Filters  \cite{karl2016deep}, Kalman Variational Autoencoders \cite{fraccaro2017disentangled}, Stochastic Recurrent Networks \cite{bayer2014learning}, and Variational Recurrent Neural Networks \cite{chung2015recurrent}. 


For nonlinear system identification, one does not only look into learning a process that is able to generate the observed variables from given inputs, but more importantly attempts to infer information on the inherent nature (dynamics) of the system. Dynamical VAEs offer a potent means to the former task, i.e., in the reconstruction of the observed signals \cite{karl2016deep}, and have been exploited in the context of nonlinear dynamical systems identification in a number of works \cite{gedon2020deep}. However, due to the generalized and highly versatile formulation of neural networks, the learned latent space often lacks a physical interpretation and is typically treated as a black-box structure. On the other hand, the availability or inference of a physically interpretable latent space is valuable across a wide range of applications, especially if these are to be used in downstream applications, as for instance for decision support \cite{sutton2018reinforcement,kratzwald2018deep,everett2018motion}.

To tackle the challenge of interpretability, we here build a learning framework for nonlinear dynamical systems by coupling a generative model with an inference model. The generative model is assembled by additively combining a deep generative model with a physics-based state space model, while the inference model adopts the structure of the original DMM, as proposed by Krishnan et al. \cite{krishnan2017structured}. This delivers a hybrid modeling framework for learning the dynamics of nonlinear systems from observations, while dealing with uncertainties relating to both the employed model and the observations/measurements. Specifically, the transition process is boosted with a physics-based term enhanced via addition of a neural network term, which aims to learn the discrepancy between the physics-based model and the actual (monitored) dynamical system. Moreover, the variance terms of the assumed transition and emission distributions are modelled as functions of the latent states and are further parameterized by a neural network in order to reflect the uncertainty in the prediction. This leads to the key point of the proposed \textit{Physics-guided Deep Markov Model} (PgDMM): by incorporating prior knowledge of the model structure, i.e., of the underlying physics, into the generative model, the framework encourages a latent space representation that aligns with the expected physics, while a neural network-based discrepancy term accounts for the knowledge that is missing.

The proposed PgDMM framework exploits the expressive power of deep learning, while retaining the driving physics of the dynamical system by partially imposing structure on the latent space. We verify the benefits of this fusion in terms of improving the predictive performance of the trained models on the simulated examples of a nonlinear dynamic pendulum and the problem of fatigue crack growth. We further validate our proposed framework on an experimental dataset, the Silverbox problem, which is broadly used as a nonlinear system identification benchmark. Our results indicate that the structured nature of the employed transition and emission functions essentially shifts the structure of the inferred latent space toward a physically interpretable formulation.

\section{Background for the Physics-guided Deep Markov Model}
\subsection{Probabilistic graphical models for Nonlinear Dynamics}\label{sec:ndyn}
\begin{figure}[H]
	\centering
    \includegraphics[width=0.7\linewidth]{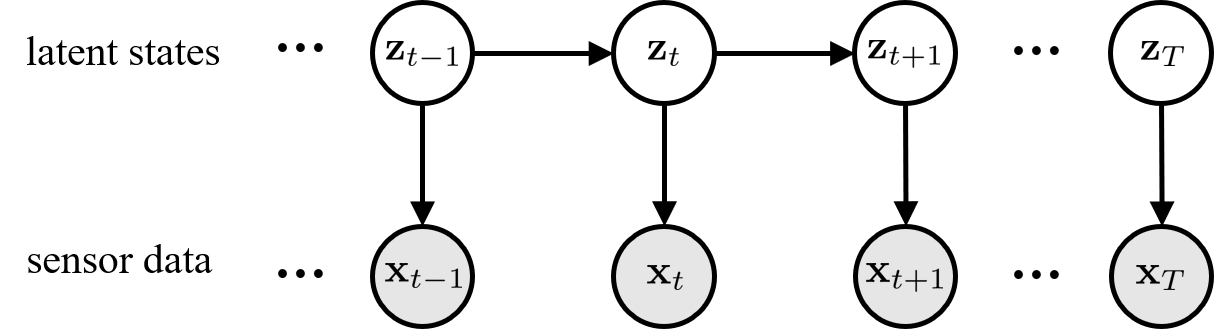}
	\caption{A structured probabilistic graphical model for nonlinear dynamical systems: $\textbf{z}_{1:T}$ is assumed to be a latent Markovian representation of the observation $\textbf{x}_{1:T}$.}
	\label{fig:graphical model}
\end{figure}
As illustrated in Figure \ref{fig:graphical model}, we consider a discretized nonlinear dynamical system or process with \textit{observations} stemming from \textit{sensor data} $\textbf{x}_{1:T}=(\textbf{x}_1,\textbf{x}_2,...,\textbf{x}_T)\subseteq\mathcal{X}\subseteq\mathbb{R}^{n_x}$ of length $T$, {\color{black} where $n_x$ is the dimension of the observation space}. It is assumed that the observation $\textbf{x}_{1:T}$ comprises a latent space representation $\textbf{z}_{1:T} = (\textbf{z}_1, \textbf{z}_2, ..., \textbf{z}_T)$, where $\textbf{z}_t\in\mathcal{Z}\subseteq\mathbb{R}^{n_z}$ is the \textit{latent variable} or \textit{latent state}, which describes the dynamics of the given system, {\color{black} and $n_z$ is the dimension of the latent space.} $\textbf{z}_t$ may also depend on \textit{external inputs} or \textit{actions} $\textbf{u}_{1:T}=(\textbf{u}_1,\textbf{u}_2,...,\textbf{u}_T)\subseteq\mathcal{U}\subseteq\mathbb{R}^{n_u}$, {\color{black} where $n_u$ is the dimension of the action space,} if the system is subjected to external forces. $\textbf{u}_{1:T}$ is not included in the following derivation for simplicity, unless otherwise noted.

One can account for the uncertainty that is inherent in such a nonlinear dynamical system, where observations stem from measurements and the model itself is not necessarily known, by adopting a structured probabilistic model, and a marginal likelihood function is defined as a probability distribution $p_\theta(\textbf{x}_{1:T})$  parameterized by $\theta\in\Theta\subseteq\mathbb{R}^{n_\theta}$, where $\theta$ designates the vector of all parameters involved in the system model {\color{black} and $n_\theta$ is the dimension of the parameter space}. With the consideration that $\textbf{x}_{1:T}$ is conditioned on $\textbf{z}_{1:T}$, the marginal likelihood function is decomposed as:
\begin{equation}\label{eq:likelihood}
p_\theta(\textbf{x}_{1:T})=\int p_{\theta_{\textbf{e}}}(\textbf{x}_{1:T}|\textbf{z}_{1:T})p_{\theta_{\textbf{t}}}(\textbf{z}_{1:T})\text{d}\textbf{z}_{1:T},
\end{equation}
in which, a \textit{transition model} $p_{\theta_{\textbf{t}}}(\textbf{z}_{1:T})$ parameterized by $\theta_{\textbf{t}}$ is considered for describing how the dynamical system (or process equation) evolves over time. Next, an \textit{emission model} $p_{\theta_{\textbf{e}}}(\textbf{x}_{1:T}|\textbf{z}_{1:T})$ parameterized by $\theta_{\textbf{e}}$ is set up for governing the relationship between the observed variables and the latent states. $\theta = \theta_{\textbf{t}} \cup \theta_\textbf{e}$ designates the vector of all parameters involved in this generative model, resulting as a concatenation of transition and emission parameters.
A nonlinear dynamical system is typically represented in the following nonlinear state-space form \cite{gedon2020deep}:
\begin{equation}
\begin{split}
\textbf{z}_t &= f_{\theta_\textbf{t}}(\textbf{z}_{t-1}, \textbf{u}_t) + \textbf{w}_t,\;\\ \textbf{x}_t &= g_{\theta_\textbf{e}}(\textbf{z}_t) + \textbf{v}_t,
\end{split}\label{ss_eq}
\end{equation}
where $\textbf{w}_t$ and $\textbf{v}_t$ are noise terms that reflect the uncertainties involved in the transition and emission processes, respectively. Motivated by this temporal nature, conditional independence assumptions on transition and emission models are further imposed, as indicated by the structured graphical model shown in Figure \ref{fig:graphical model}, and they are decomposed as:
\begin{subequations}\label{trans_emi}
\begin{equation}\label{transition}
    p_{\theta_\textbf{t}}(\textbf{z}_{1:T})=\prod_{t=1}^Tp_{\theta_\textbf{t}}(\textbf{z}_{t}|\textbf{z}_{t-1}), \qquad\quad \text{\textit{transition model}} \\
\end{equation}
\begin{equation}\label{emission}  
    p_{\theta_\textbf{e}}(\textbf{x}_{1:T}|\textbf{z}_{1:T})=\prod_{t=1}^Tp_{\theta_\textbf{e}}(\textbf{x}_{t}|\textbf{z}_{t}), \qquad \text{\textit{emission model}}
\end{equation}
\end{subequations}
Eq.\eqref{transition} assumes that $\mathbf{z}_{1:T}$ is Markovian, i.e., that the current state $\textbf{z}_{t}$ only depends on the previous state $\mathbf{z}_{t-1}$, since a Markov process is memoryless. Eq.\eqref{emission} implies that the current observation $\textbf{x}_{t}$ only depends on the current state $\textbf{z}_{t}$. It is also noted that in Eq.\eqref{trans_emi}, throughout the sequence (from $t = 1$  to $t = T$), the transition and emission models do not change over time, i.e., $\theta_\textbf{t}$ and $\theta_\textbf{e}$ are independent of time $t$.

The transition and emission models can be flexibly constructed; for example, $\theta_\textbf{t}$ and $\theta_\textbf{e}$ can be parameters involved in purely physics-based transition and emission models \cite{lund2020variational} or, alternatively, $\theta_\textbf{t}$ and $\theta_\textbf{e}$ can further correspond to neural networks parameters, as is the case for deep generative models \cite{rezende2014stochastic}.  

\subsection{Learning nonlinear dynamical systems with variational inference}
The process of learning a dynamical system from a given training dataset $\textbf{x}_{1:T}$ can be mathematically formalized as the determination of the vector of parameters $\theta$ that are involved in the transition and emission models. This is accomplished by maximizing the marginal log-likelihood function $\log p_\theta(\textbf{x}_{1:T})$ defined in Eq.\eqref{eq:likelihood}, following the principle of maximum likelihood estimation.
The marginal log-likelihood function is further expressed as follows (the subscripts of $\textbf{x}_{1:T}$ and $\textbf{z}_{1:T}$ are dropped here for simplicity):
\begin{equation}
\begin{split}
\log p_\theta(\textbf{x}) & =  \mathbb{E}_{\textbf{z}\sim p_\theta(\textbf{z}|\textbf{x})}[\log p_\theta(\textbf{x})]\quad \\
&=\mathbb{E}_{\textbf{z}\sim p_\theta(\textbf{z}|\textbf{x})}[\log p_\theta(\textbf{x},\textbf{z}) - \log p_\theta(\textbf{z}|\textbf{x})]
\end{split}
\end{equation}
Note that the first equality holds due to the fact that $\log p_\theta(\textbf{x})$ is deterministic with respect to $p_\theta(\textbf{z}|\textbf{x})$. The posterior $p_\theta(\textbf{z}|\textbf{x})$ corresponds to the distribution over the latent variables given the observed data. The computation of this quantity is often required for implementing a learning rule. 
\textit{Variational Inference}\cite{kingma2013auto}, 
introduced very recently, seeks to derive a distribution $q_\phi(\textbf{z}|\textbf{x})$ that is parameterized by neural networks, for approximating the intractable true posterior distribution $p_\theta(\textbf{z}|\textbf{x})$. The \textit{evidence lower bound} (ELBO) of the marginal log-likelihood $\log p_\theta(\textbf{x})$ is adopted as the loss function $\mathcal{L}(\theta,\phi;\textbf{x})$ to be maximized. Its form is yielded by Jensen's inequality, which states that $g(\mathbb{E}[\textbf{X}])\geq\mathbb{E}[g(\textbf{X})]$ for a concave function $g$ and a random variable $\textbf{X}$ and is often used to bound the expectation of a convex/concave function:
\begin{equation}\label{ELBO}
\begin{split}
\log p_\theta(\textbf{x})&=\log\int p_\theta(\textbf{x},\textbf{z})d\textbf{z}=\log\mathbb{E}_{q_\phi(\textbf{z}|\textbf{x})}\Big[\frac{p_\theta(\textbf{x},\textbf{z})}{q_\phi(\textbf{z}|\textbf{x})}\Big]\\
&\geq\mathbb{E}_{q_\phi(\textbf{z}|\textbf{x})}\Big[\log\frac{p_\theta(\textbf{x},\textbf{z})}{q_\phi(\textbf{z}|\textbf{x})}\Big]=\mathbb{E}_{q_\phi(\textbf{z}|\textbf{x})}[\log p_\theta(\textbf{x},\textbf{z})-\log q_\phi(\textbf{z}|\textbf{x})]\\
&=\underbrace{\mathbb{E}_{q_\phi(\textbf{z}|\textbf{x})}[\log p_{\theta_e}(\textbf{x}|\textbf{z})]}_{\text{reconstruction}}-\underbrace{\text{KL}(q_\phi(\textbf{z}|\textbf{x})||p_{\theta_t}(\textbf{z}))}_{\text{regularization}}=:\mathcal{L}(\theta,\phi;\textbf{x}),
\end{split}
\end{equation}
where $\text{KL}(\cdot||\cdot)$ is defined as $\text{KL}[q(\textbf{z}|\textbf{x})||p(\textbf{z})] := \int q(\textbf{z}|\textbf{x}) \log\frac{q(\textbf{z}|\textbf{x})}{p(\textbf{z})} d\textbf{z}$, the Kullback–Leibler (KL) divergence. The loss function comprises a reconstruction term and a regularization term, where the former evaluates the accuracy of the chained process of encoding-decoding and the latter penalizes the loss, enforcing the closeness between $q_\phi(\textbf{z}|\textbf{x})$ and $p_\theta(\textbf{z}|\textbf{x})$. It is noted that the temporal dependence introduced in Eq.\eqref{trans_emi} can be applied to the loss function (the detailed derivation is provided in Appendix \ref{app:elbo}) and is further derived as:
\begin{equation}\label{ELBO_factorized}
\begin{split}
\mathcal{L}(\theta,\phi;\textbf{x})
&= \sum_{t=1}^{T}\mathbb{E}_{q_\phi}[\log p_{\theta_e}(\textbf{x}_t|\textbf{z}_t)-\text{KL}(q_\phi(\textbf{z}_t|\textbf{z}_{t-1},\textbf{x})||p_{\theta_t}(\textbf{z}_t|\textbf{z}_{t-1}))] \\
&= \sum_{t=1}^{T}\mathbb{E}_{q_{\phi}}[\log \underbrace{p_{\theta_{\textbf{t}}}(\textbf{z}_t|\textbf{z}_{t-1})}_{\text{transition}} + \log \underbrace{p_{\theta_{\textbf{e}}}(\textbf{x}_t|\textbf{z}_t)}_{\text{emission}} - \log \underbrace{q_{\phi}(\textbf{z}_t|\textbf{z}_{t-1},\textbf{x})}_{\text{inference}} ]
\end{split}
\end{equation}
where the loss function is further interpreted as the summation of three terms related to transition, emission, and inference. The objective lies in maximizing the ELBO $\mathcal{L}(\theta,\phi;\textbf{x})$ by computing the gradient with respect to both $\theta$ and $\phi$:
\begin{align}\label{gradient}
\sum_{t=1}^{T}\nabla_{\theta, \phi}\mathbb{E}_{q_{\phi}}[\log p_{\theta_{\textbf{t}}}(\textbf{z}_t|\textbf{z}_{t-1}) + \log p_{\theta_{\textbf{e}}}(\textbf{x}_t|\textbf{z}_t) - \log q_{\phi}(\textbf{z}_t|\textbf{z}_{t-1},\textbf{x}) ]
\end{align}
Since the terms in Eq.\eqref{gradient} cannot typically be integrated in closed form, the gradients are usually practically estimated by extracting $S$ samples from $q_\phi(\textbf{z}|\textbf{x})$ via use of Markov chain Monte Carlo (MCMC) and computing, e.g.,
\begin{equation}\label{MCMC}
\nabla_{\theta_\textbf{t}}\mathbb{E}_{q_\phi}[\log p_{\theta_{\textbf{t}}}(\textbf{z}_t|\textbf{z}_{t-1})] \approx \frac{1}{S}\sum_{s=1}^{S}\nabla_{\theta_\textbf{t}}\log p_{\theta_{\textbf{t}}}(\textbf{z}_t^{(s)}|\textbf{z}_{t-1}^{(s)}),
\end{equation}
where $\textbf{z}_t^{(s)}$ is a sample of the latent state.
{\color{black} When all the involved distributions are Gaussian, the KL-divergence terms comprise an analytical form and the loss ELBO (\ref{ELBO_factorized}) can be computed more explicitly as:
\begin{equation}\label{ELBO_analytic}
\begin{split}
\mathcal{L}(\theta,\phi;\textbf{x})
&= \sum_{t=1}^{T}\mathbb{E}_{q_\phi}[\log p_{\theta_e}(\textbf{x}_t|\textbf{z}_t)-\text{KL}(q_\phi(\textbf{z}_t|\textbf{z}_{t-1},\textbf{x})||p_{\theta_t}(\textbf{z}_t|\textbf{z}_{t-1}))] \\
& = -\frac{1}{2}\sum_{t=1}^{T}\mathbb{E}_{q_\phi}[\log|\Sigma_t|+(\textbf{x}_t-\mu_t)^T\Sigma_t^{-1}(\textbf{x}_t-\mu_t)+n_x\log(2\pi) \\
&+\log\frac{|\Sigma_{pt}|}{|\Sigma_{qt}|}-n_z+\text{Tr}(\Sigma_{pt}^{-1}\Sigma_{qt})+(\mu_{pt}-\mu_{qt})^T\Sigma_{pt}^{-1}(\mu_{pt}-\mu_{qt})],
\end{split}
\end{equation}
where $n_x$, $n_z$ are the respective dimensions of the observations and latent states.}


\newpage
\section{Physics-guided Deep Markov Model}\label{sec:PgDMM}
As discussed in Section \ref{sec:ndyn}, a nonlinear dynamical system can often be cast in a state-space representation and, thus, naturally comprises a latent Markovian representation, as elaborated in Eq.\eqref{ss_eq}, where  $\textbf{z}_t$ only depends on $\textbf{z}_{t-1}$ and $\textbf{x}_t$ only depends on $\textbf{z}_t$. In real-world applications, it is often non-trivial to parameterize a nonlinear physics-based model in terms of its transition and emission functions. The Deep Markov Model\cite{krishnan2017structured} relaxes this restriction by parameterizing the transition and emission functions via feed-forward neural networks. However, due to the highly versatile, and thus generic, formulation of neural network functions, the inferred latent space often lacks physical interpretation and a structured representation \cite{karl2016deep}.

\begin{figure}[H]
	\centering
    \includegraphics[width=1.0\linewidth]{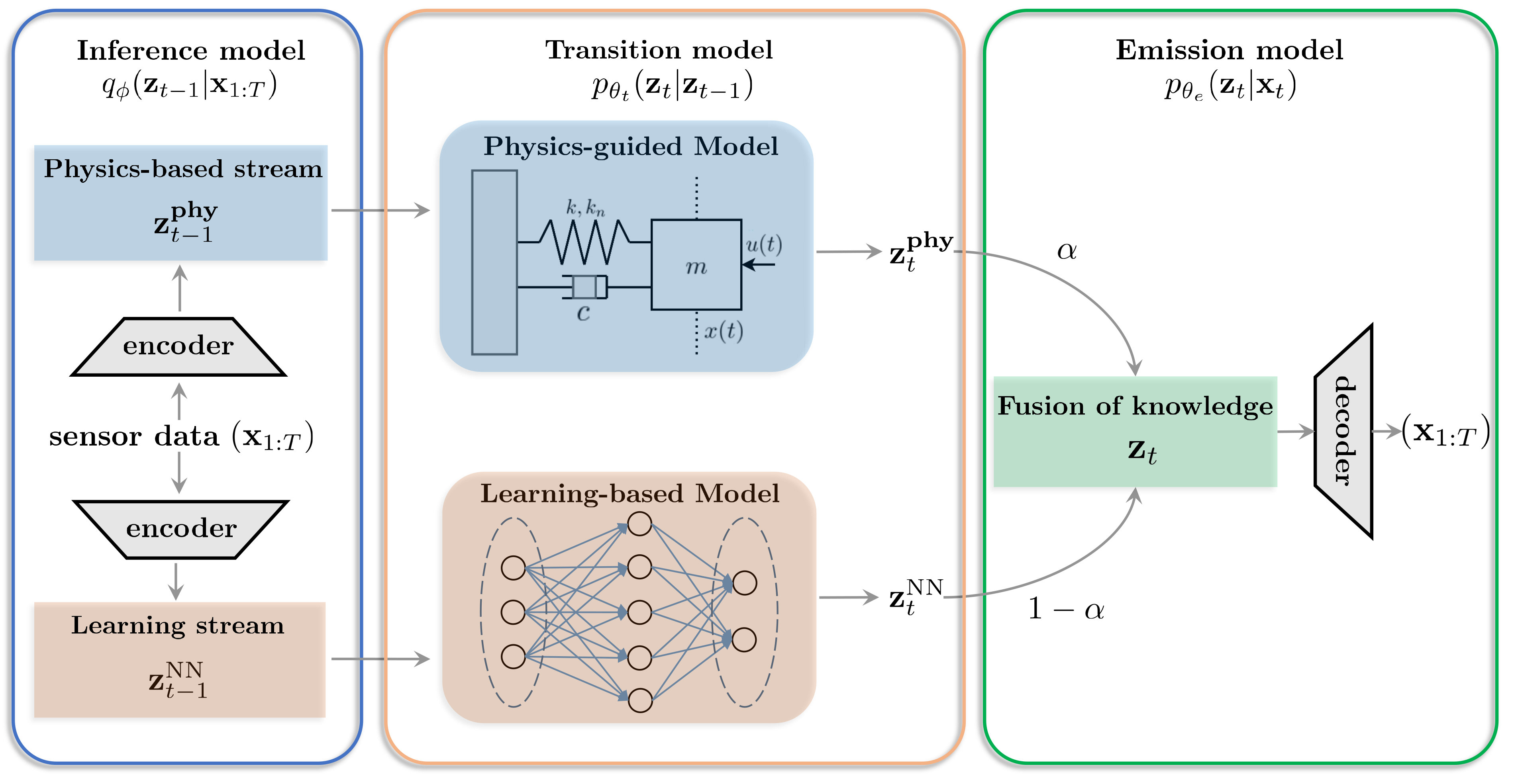}
    \vspace*{3mm}
	\caption{General learning framework of PgDMM: The latent states $\textbf{z}_{1:T}^\textbf{phy}$ and $\textbf{z}_{1:T}^\text{NN}$ are sampled from the inference networks $q^\textbf{phy}_{\phi}(\textbf{z}^\textbf{phy}_{1:T}|\textbf{x})$ and $q^\text{NN}_{\phi}(\textbf{z}^\text{NN}_{1:T}|\textbf{x})$ respectively. {\color{black} These coupled inference networks serve as the encoder, i.e., translate the observations $\textbf{x}$ into a useful latent representation (latent state) $\textbf{z}$.} $\textbf{z}_{t-1}^\textbf{phy}$ and $\textbf{z}_{t-1}^\text{NN}$ are jointly fed through the physics-based transition model (prior knowledge) and learning-based model (learned knowledge) to  update the latent states, and estimate the ELBO, i.e., the likelihood of the transition, emission, and inference processes, based on the samples, as shown in Eq.\eqref{ELBO2}. Gradients of ELBO are further estimated via MCMC, according to Eq.\eqref{MCMC}, and used to update parameters $\theta$ and $\phi$.}
	\label{fig:graphical_abstract}
\end{figure}
In this paper, as illustrated in Figure \ref{fig:graphical_abstract}, we build a learning framework for nonlinear dynamics, where the generative model (transition and emission models) is built by fusing a deep generative model and a physics-guided model and the inference model adopts the structure suggested in (Krishnan et al., 2015)\cite{krishnan2015deep}. This structure, which we term the \textit{Physics-guided Deep Markov Model} (PgDMM), assimilates possible prior knowledge on the system into the developed generative model, in order to partially guide the inference of the latent variables according to the suspected physics of the dynamical system.

\subsection{Transition Model}
\begin{figure}[H]
	\centering
    \includegraphics[width=0.82\linewidth]{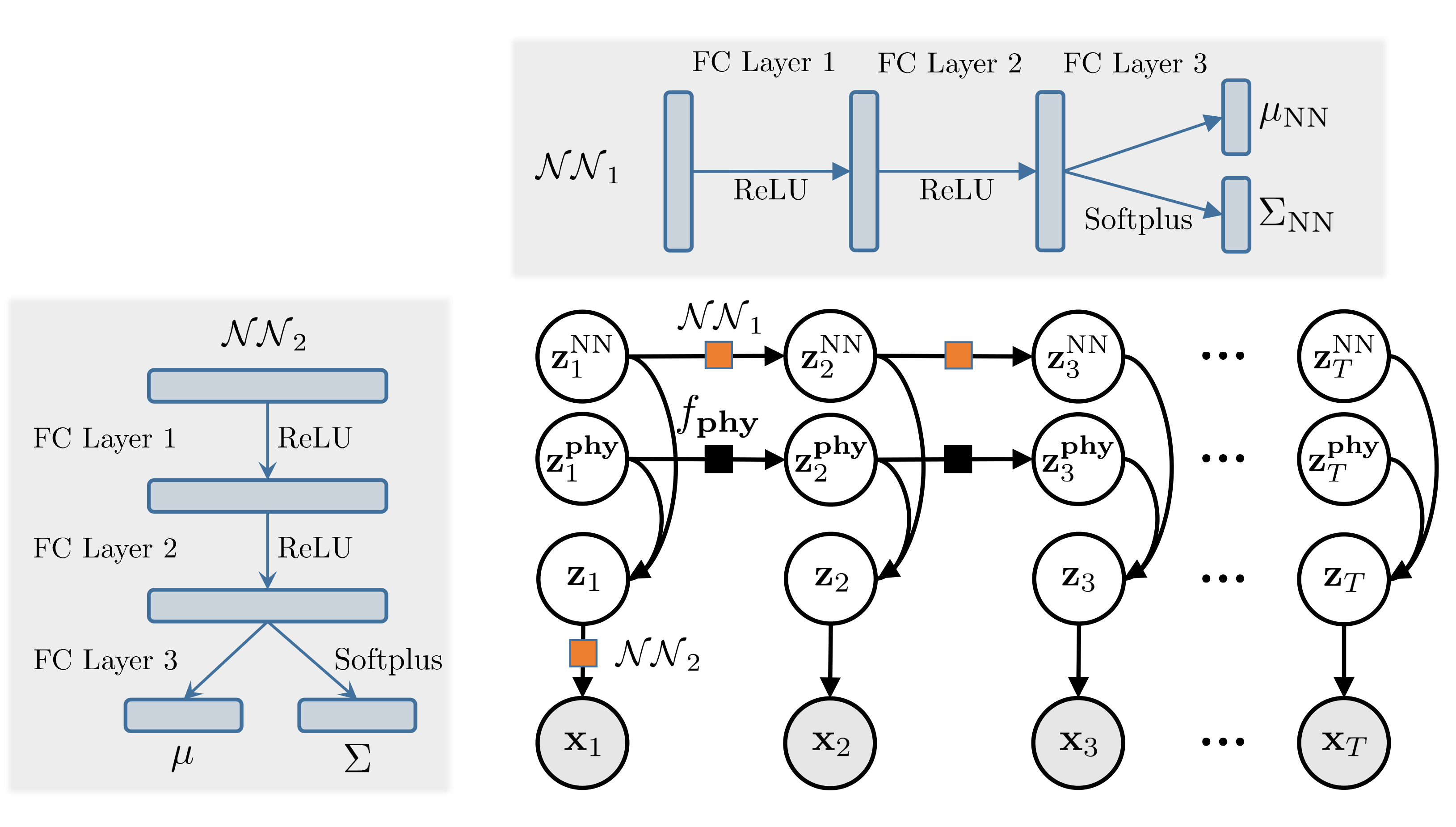}
	\caption{Illustration of the framework with modeling of transition and emission processes: the transition process is composed of a physics-guided stream $\textbf{z}_{1:T}^\textbf{phy}$ modelled by a physics-guided transition function $f_\textbf{phy}$, and a learning-based stream $\textbf{z}_{1:T}^\text{NN}$ with a transition function parameterized by a neural network $\mathcal{NN}_1$, which aims at learning the discrepancy between the physics-guided stream $\textbf{z}_{1:T}^\textbf{phy}$ and the actual dynamical system. The latent state results as a weighted combination of the two streams. The emission function is finally parameterized by another neural network $\mathcal{NN}_2$. {\color{black} The zoomed-in architectures of $\mathcal{NN}_1$ and $\mathcal{NN}_2$ are further shown in the shaded boxes and all the details are provided in Appendix \ref{app:arc}.}}
	\label{fig:model}
\end{figure} 
The transition from state $\textbf{z}_{t-1}$ to state at the next time instant $t$,  $\textbf{z}_{t}$, is formulated as a hybrid structure comprising a physics-guided model $f_{\textbf{phy}}$, and a learning-based model, which is parameterized by means of a feed-forward neural network, $\mathcal{NN}_1$.
More specifically, it is assumed that each updated state $\textbf{z}_{t}$ occurs as a linear combination of two random variables, namely $\textbf{z}_{t}^\textbf{phy}$ and $\textbf{z}_{t}^\text{NN}$, that are weighted by a factor $\alpha$ ($0\leq\alpha\leq 1$):
\begin{subequations}
    \begin{equation}\label{eq:weight}
      \textbf{z}_{t}=\alpha\textbf{z}_{t}^\textbf{phy}+(1-\alpha)\textbf{z}_{t}^\text{NN},
    \end{equation}
where $\alpha$ is set as a trainable parameter in the proposed framework, i.e., $\alpha$ is included in the parameter vector $\theta$, which was specified in Eq.\eqref{eq:likelihood}. The latent state $\textbf{z}_{t}^\textbf{phy}$ is conditioned on $\textbf{z}_{t-1}^\textbf{phy}$, following a Gaussian distribution, where the mean value is updated according to a fixed physics-guided equation $f_{\textbf{phy}}(\textbf{z}_{t-1}^\textbf{phy})$, while the variance updating is accomplished by a trainable neural network $\mathcal{NN}_0(\textbf{z}_{t-1}^\textbf{phy})$, i.e., a function of $\textbf{z}_{t-1}^{\textbf{phy}}$, since variance (uncertainty) is often not characterized by the underlying physics model: 
\begin{equation}
    \textbf{z}_{t}^\textbf{phy}\sim \mathcal{N}(\mathbf{\mu}_\textbf{phy}(\textbf{z}_{t-1}^\textbf{phy}),  \mathbf{\Sigma}_\textbf{phy}(\textbf{z}_{t-1}^\textbf{phy})),
\end{equation}
where
\begin{equation}
\begin{split}
& \mathbf{\mu}_\textbf{phy}(\textbf{z}_{t-1}^\textbf{phy}) = f_{\textbf{phy}}(\textbf{z}_{t-1}^\textbf{phy}), \\
& \mathbf{\Sigma}_\textbf{phy}(\textbf{z}_{t-1}^\textbf{phy}) =  \mathcal{NN}_0(\textbf{z}_{t-1}^\textbf{phy}).
\end{split}
\end{equation}

Similarly, $\textbf{z}_{t}^\text{NN}$ is conditioned on $\textbf{z}_{t-1}^\text{NN}$, also following a Gaussian distribution with the corresponding mean $\mathbb{\mu}_\text{NN}(\textbf{z}_{t-1}^\text{NN})$ and variance $\mathbf{\Sigma}_\text{NN}(\textbf{z}_{t-1}^\text{NN})$ computed as the two outputs of a trainable neural network $\mathcal{NN}_1$:
\begin{equation}
    \textbf{z}_{t}^\text{NN}\sim \mathcal{N}(\mathbf{\mu}_\text{NN}(\textbf{z}_{t-1}^\text{NN}),\mathbf{\Sigma}_\text{NN}(\textbf{z}_{t-1}^\text{NN})),
\end{equation}
in which (also illustrated in the shaded box in Figure \ref{fig:model}),
\begin{equation}\label{Eq:NN1}
   [\mathbf{\mu}_\text{NN}(\textbf{z}_{t-1}^\text{NN}),\; \mathbf{\Sigma}_\text{NN}(\textbf{z}_{t-1}^\text{NN})] = \mathcal{NN}_1(\textbf{z}_{t-1}^\text{NN}).
\end{equation}
\end{subequations}
Note that $\textbf{z}_t$ is a random variable, thus  Eq.\eqref{eq:weight} implies that
\begin{equation}\label{mean_var}
\begin{split}
\mathbf{\mu}(\textbf{z}_t)&=\alpha\mathbf{\mu}_\textbf{phy}(\textbf{z}_{t-1}^\textbf{phy})+(1-\alpha)\mathbf{\mu}_\text{NN}(\textbf{z}_{t-1}^\text{NN}),\\
\mathbf{\Sigma}(\textbf{z}_t)&=\alpha^2\mathbf{\Sigma}_\textbf{phy}(\textbf{z}_{t-1}^\textbf{phy})+(1-\alpha)^2\mathbf{\Sigma}_\text{NN}(\textbf{z}_{t-1}^\text{NN}).
\end{split}
\end{equation}

For the first random variable $\textbf{z}_t^\textbf{phy}$, the physics-guided (or model-based) transition function $f_\textbf{phy}( \textbf{z}_t^\textbf{phy})$ is formulated based on assumed prior knowledge on the system response, i.e., by adoption of a structured model for part of the transition process. It is noted that $f_\textbf{phy}$ does not have to fully represent the actual transition process, but aims to only account for some partial knowledge of the system. A typical example for the case of nonlinear dynamical system modeling is understood for instance in the use of a linear approximation function of $\textbf{z}_t^\textbf{phy}$ when the system comprises nonlinear dynamics. The covariance matrices $\mathbf{\Sigma}_\text{\textbf{phy}}$ and $\mathbf{\Sigma}_\text{NN}$ are assumed to be diagonal. 
This is a sound approximation in most cases, as the uncertainty can be assumed to be adequately described by a white noise assumption, whose noise to signal ratio is low.

The second random variable $\textbf{z}_t^\text{NN}$, which is parameterized by a neural network, intends to learn the discrepancy between the approximation offered by $\textbf{z}_t^\textbf{phy}$ and the actual system dynamics. $\textbf{z}_t^\text{NN}$ is guided by the introduction of $f_{\textbf{phy}}$ in maximizing the likelihood, which implies that the training is less prone to learning arbitrary functions that simply aim to fit the observations.
\subsection{Emission Model}\label{sec:emission}
The emission process, which guides the transition from the latent state $\textbf{z}_t$ to the observation $\textbf{x}_{t}$ reads:
\begin{equation}
\textbf{x}_t\sim P(E(\textbf{z}_t)),
\end{equation}
where $P$ is the probability distribution of measured data $\textbf{x}_t$ that can be parameterized by an additional neural network term $\mathcal{NN}_2$, which is used to parameterize a suitable distribution, depending on the application at hand. For example, if $P$ is a Gaussian distribution, then in analogy to Eq.\eqref{Eq:NN1}, the neural network will be used to parameterize the mean and variance parameters $E(\textbf{z}_t) = [\mu(\textbf{z}_t), \mathbf{\Sigma}(\textbf{z}_t)] = \mathcal{NN}_2(\textbf{z}_t)$, while if $P$ is a Bernoulli distribution, then the parameterization is defined as $E(\textbf{z}_t) = p(\textbf{z}_t) = \mathcal{NN}_2(\textbf{z}_t)$, where $p$ is the parameter of the Bernoulli distribution.

\subsection{Inference Model}
\begin{figure}[h]
	\centering
    \includegraphics[width=0.61\linewidth]{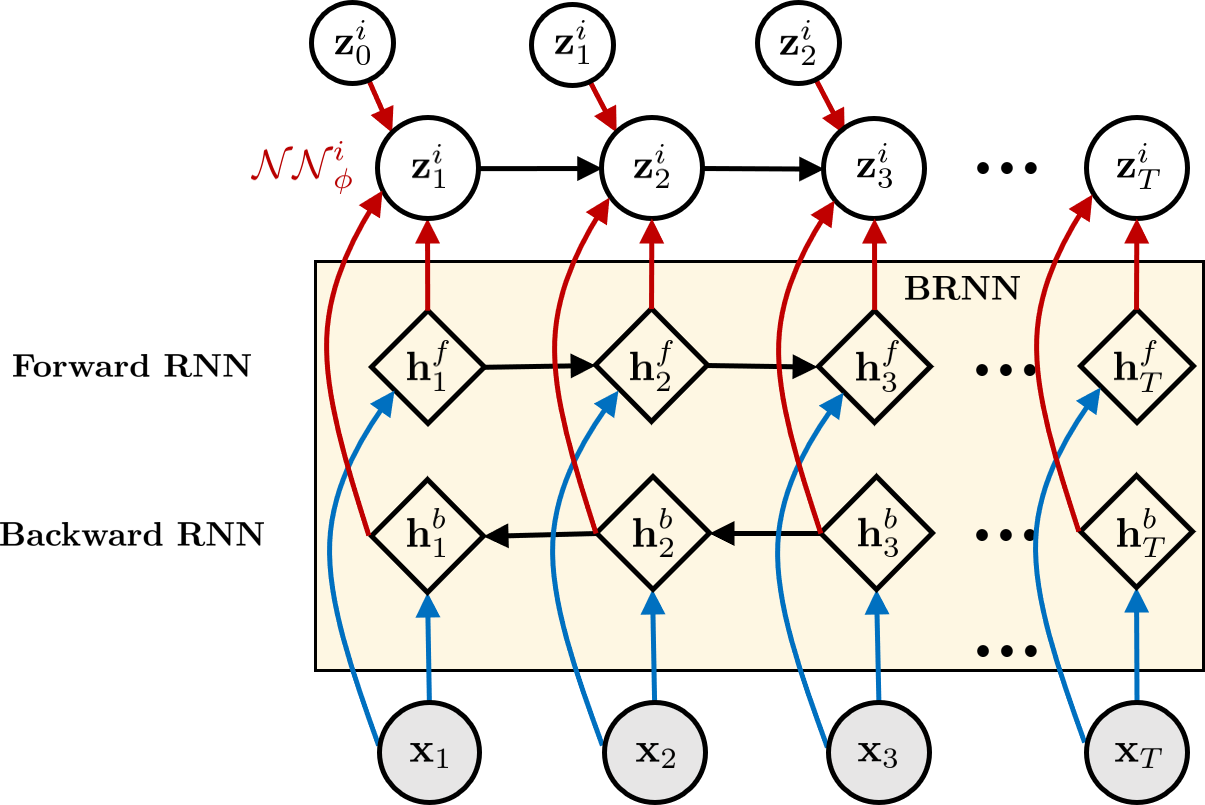}
    \vspace*{3mm}
	\caption{Inference network for PgDMM: The approximate inference model $q_\phi^i(\textbf{z}_{1:T}^i|\textbf{x}_{1:T})$ for $\textbf{z}_{1:T}^i$, where $i$ stands for either \textbf{phy} or \text{NN} since the structure of both inference models is identical, is derived using bidirectional recurrent neural networks (BRNNs). The BRNN admits $\textbf{x}_{1:T}$ as inputs and - through a series of hidden layers of the neural network, denoted by \textcolor{blue}{blue} arrows - forms two sequences of deterministic hidden states $\textbf{h}_t^\text{f}$ and $\textbf{h}_t^\text{b}$. These express the flow of information in the forward (from left to right) and backward (from right to left) sense, respectively. These, together with the previous latent state $\textbf{z}_{t-1}^i$, are fed into a further neural network $\mathcal{NN}_\phi^i$, denoted by the \textcolor{red}{red} arrows. Finally, the inference network outputs 
	the mean and covariance of $q^i_{\phi}(\textbf{z}^i_{t}|\textbf{z}^i_{t-1},\textbf{x})$, as shown in Eq.\eqref{eq:infer}.}
	\label{fig:inference}
\end{figure}
The true posterior distribution $p(\textbf{z}_{1:T}|\textbf{x}_{1:T})$, which reflects the joint distribution of latent variables $\textbf{z}_{1:T}$ for observations up to time $T$, $\textbf{x}_{1:T}$, is intractable in the context of Deep Markov modeling. It can be approximated by a distribution $q_\phi(\textbf{z}_{1:T}|\textbf{x}_{1:T}) \approx p(\textbf{z}_{1:T}|\textbf{x}_{1:T})$\cite{kingma2019introduction}, forming an inference model, as illustrated in Figure \ref{fig:inference}. The inference network follows the structure suggested in the original DMM scheme \cite{krishnan2017structured}.
 
Given the independence statements implied by the graphical model in Figure \ref{fig:graphical model}, the latent variables $\textbf{z}_{1:T}$ retain Markovian behavior when conditioning on $\textbf{x}_{1:T}$ and the true posterior $p(\textbf{z}|\textbf{x})$ can be factorized as:
\begin{equation}
p(\textbf{z}|\textbf{x})=p(\textbf{z}_0)\prod_{t=1}^Tp(\textbf{z}_t|\textbf{z}_{t-1},\textbf{x}),
\end{equation}
where $\textbf{z}$ is used as short for $\textbf{z}_{1:T}$ and $\textbf{x}$ is short for $\textbf{x}_{1:T}$ and $\textbf{z}_0$ is a pseudo node representing the initial condition (trainable), i.e., $p(\textbf{z}_t|\textbf{z}_{t-1},\textbf{x})$ is reduced to $p(\textbf{z}_0)$ when $t=0$. The approximated distribution $q_\phi(\textbf{z}|\textbf{x})$ is designed to be a Gaussian distribution that mimics the behavior of the true posterior. Note that, as there exist two streams in the transition model, namely the physics-guided ($\textbf{z}_t^\textbf{phy}$) and learning-based stream ($\textbf{z}_t^\text{NN}$), the final combined $q_\phi(\textbf{z}|\textbf{x})$ is calculated as:
\begin{equation}
\begin{split}
q_\phi(\textbf{z}|\textbf{x}) & = q_\phi^{\textbf{phy}}(\textbf{z}^\textbf{phy}|\textbf{x})q_\phi^{\text{NN}}(\textbf{z}^\text{NN}|\textbf{x})\\
& = q_\phi^\textbf{phy}(\textbf{z}^\textbf{phy}_0)q_\phi^{\text{NN}}(\textbf{z}^\text{NN}_0)\prod_{t=1}^Tq_\phi^\textbf{phy}(\textbf{z}^\textbf{phy}_t|\textbf{z}^\textbf{phy}_{t-1},\textbf{x})\prod_{t=1}^Tq_\phi^{\text{NN}}(\textbf{z}^\text{NN}_t|\textbf{z}^\text{NN}_{t-1},\textbf{x}), \\
&(\text{with Markovian property})
\end{split}
\end{equation}
and the two terms are further obtained as:
\begin{subequations}
\begin{equation}\label{eq:infer}
q^i_{\phi}(\textbf{z}^i_{t}|\textbf{z}^i_{t-1},\textbf{x}) = 
\mathcal{N}(\mathbf{\mu}_{\phi}^i(\textbf{z}^i_{t-1},\textbf{x}),\; \mathbf{\Sigma}_{\phi}^i(\textbf{z}^i_{t-1},\textbf{x})),  \quad\quad (\text{$i = \textbf{phy}$ or NN})
\end{equation}
where the superscript $i$ stands for either \textbf{phy} or \text{NN}, representing the two information streams that were introduced in Eq.\eqref{eq:weight}. $\mathbf{\mu}_\phi^i$ and $\mathbf{\Sigma}_\phi^i$ designate the mean values and covariance matrix of the distribution, respectively, given by the outputs of neural networks $\mathcal{NN}^i_\phi$:
\begin{equation}\label{nn_phi}
   [\mathbf{\mu}_{\phi}^i(\textbf{z}^i_{t-1},\textbf{x}),\; \mathbf{\Sigma}_{\phi}^i(\textbf{z}^i_{t-1},\textbf{x})] = \mathcal{NN}_\phi^i(\textbf{h}_{t}^i)  , 
\end{equation}
where
\begin{equation}
  \textbf{h}_t^i = \frac{1}{3}[\textbf{h}_t^b+\textbf{h}_t^f + \tanh(\textbf{W}^i\textbf{z}_{t-1}^i+\textbf{b}^i)].
  \label{eq:comb}
\end{equation}
\end{subequations}
{\color{black} The right hand side of Eq.\eqref{eq:comb} is termed as the combiner function and the hyperbolic tangent tanh is used as the activation function, which is charged with modeling the possibly nonlinear term of $\textbf{z}_{t-1}^i$ in the inference network \cite{krishnan2017structured}.} In this step, as further illustrated in Figure \ref{fig:inference}, the complete sequence of the observations $\textbf{x}_{1:T}$ is processed by a bidirectional recurrent neural network (BRNN)\cite{schuster1997bidirectional}. The BRNN accepts the sequence $\textbf{x}_{1:T}$ as input and outputs two sequences of hidden states $\textbf{h}_t^\text{f}$ and $\textbf{h}_t^\text{b}$, representing information flow forward (from left to right) and backward (from right to left) respectively. The past and future observations encoded in $\textbf{h}_t^\text{f}$ and $\textbf{h}_t^\text{b}$ are combined with the last state $\textbf{z}_{t-1}^i$, as illustrated Eq.\eqref{eq:comb}. A single layer feed-forward neural network (parameterized by $\textbf{W}^i$ and $\textbf{b}^i$) is adopted to transform $\textbf{z}_{t-1}^i$ into the same dimension as the hidden state $\textbf{h}_t^b$ and $\textbf{h}_t^f$. 

Note that in Eq.\eqref{eq:infer}, although the inference of $\textbf{z}^\textbf{phy}$ from $q_\phi^\textbf{phy}$ and of $\textbf{z}^\text{NN}$ from $q_\phi^\text{NN}$ is conducted individually, the same bidirectional neural network is applied to both streams, which implies that the parameters $\phi$ are shared (the weights in the BRNN), i.e., a shared function is used to learn the dependence of the posterior distribution $q^i_{\phi}(\textbf{z}^i_{t}|\textbf{z}^i_{t-1},\textbf{x}_{1:T})$ on $\textbf{x}_{1:T}$. {\color{black} As a consequence, the hidden states $\textbf{h}_t^b$,  $\textbf{h}_t^f$ of the BRNN are the same for both inference networks; the main difference of these inference networks lies in the weights $\textbf{W}^i$, $\textbf{b}^i$ of the combiner function, as shown in Eq.\eqref{eq:comb} and $\mathcal{NN}_\phi^i$.}

\subsection{The Evidence Lower Bound Objective Function (ELBO) of PgDMM}
Similar to the derivation of an ELBO, given in Eq.\eqref{ELBO}, a lower bound can be obtained for PgDMM (see details in Appendix \ref{app:ELBO}):
\begin{equation}\label{ELBO2}
\begin{split}
\log p(\textbf{x})&\geq\mathcal{L}_\textbf{phy}(\theta,\phi;\textbf{x})\\
&:=\mathbb{E}_{q_\phi}[\log p_\theta(\textbf{x}|\textbf{z})]-\big\{\text{KL}[q_\phi^{\textbf{phy}}(\textbf{z}^\textbf{phy}|\textbf{x})||p(\textbf{z}^\textbf{phy})]+\text{KL}[q_\phi^{\text{NN}}(\textbf{z}^\text{NN}|\textbf{x})||p(\textbf{z}^\text{NN})]\big\} \\
&=\sum_{t=1}^{T}\mathbb{E}_{q_\phi}\log p(\textbf{x}_t|\textbf{z}_t) \\
&-\sum_{t=1}^T\mathbb{E}_{q_\phi^\textbf{phy}(\textbf{z}_{t-1}^\textbf{phy}|\textbf{x}_{1:T})}[\text{KL}(q_\phi^\textbf{phy}(\textbf{z}_t^\textbf{phy}|\textbf{z}_{t-1}^\textbf{phy},\textbf{x}_{1:T})||p_\theta(\textbf{z}_t^\textbf{phy}|\textbf{z}_{t-1}^\textbf{phy}))] \\
&-\sum_{t=1}^T\mathbb{E}_{q_\phi^\text{NN}(\textbf{z}_{t-1}^\text{NN}|\textbf{x}_{1:T})}[\text{KL}(q_\phi^\text{NN}(\textbf{z}_t^\text{NN}|\textbf{z}_{t-1}^\text{NN},\textbf{x}_{1:T})||p_\theta(\textbf{z}_t^\text{NN}|\textbf{z}_{t-1}^\text{NN}))].
\end{split}
\end{equation}
Eq.\eqref{ELBO2} implies that the loss function encodes the trade-off between reconstruction accuracy (model fit), via the expected log-likelihood term, and the prior regularization, via the KL-divergence term. The introduction of a physics-guided term into the model structure is equivalent to splitting the prior regularization into two terms, where the former acts as a physical constraint. For the variational inference implementation, we rely heavily on the automated processes provided by the Pyro library\cite{bingham2018pyro}. We let the library handle the computation of the ELBO as well as implement all Monte Carlo simulations necessary for stochastic backpropagation. {\color{black} The data and codes used in this paper are publicly available on GitHub at https://github.com/liouvill/PgDMM.}



\section{Simulation and Experiment Results}\label{exp}
In this section, we investigate the proposed hybrid modeling scheme, the PgDMM, in terms of its performance in recovering a latent representation with physical connotation on the basis of observations from nonlinear dynamical systems. The following models are adopted here as case studies for this investigation: (i) a pendulum system, where particularly the observations are image sequences (snapshots of the pendulum motions); (ii) a fatigue crack growth problem, and (iii) the Silverbox nonlinear benchmark \cite{wigren2013three} with real-world data. The first example attempts to demonstrate the capacity of PgDMM in learning from high-dimensional {\color{black} image} data and modeling complex emission processes. The second example aims to show the potential in uncertainty qualification with known ground-truth uncertainty, while the last experiment serves as an application onto a more complex experimental case study. {\color{black} At this point, it is important to clarify that, in all the following comparison between PgDMM and the original DMM, the latent states of the DMM are learned without inclusion of the physics-guided component in any part of the learning and prediction process. This is equivalent to use of the original DMM framework, where the learning-based component is charged with modeling the complete latent state. To the contrary, in the PgDMM the data-driven component $\textbf{z}^\text{NN}_t$, is used to only model the discrepancy between the physics-guided component $\textbf{z}^\textbf{phy}_t$ and the full latent states $\textbf{z}_t$.}

\subsection{Dynamic Pendulum}
{\color{black} We adopt image data as the observation source for this example, in order to demonstrate applicability of our work to learning from image/video information. Optical measurements are increasingly {\color{black} gaining ground} in the SHM domain, where vision-based information becomes more and more available {\color{black} even for monitoring of dynamical systems \cite{harmanci2019novel,yang2017blind}}. Image data are intrinsically high-dimensional and much recent work has focused on learning patterns from such data sources. It represents a setting, where the posterior is complex and often requires highly nonlinear models for its approximation. Thus, the formulation of adequate emission models can be particularly difficult in this case. Conventional VAEs have traditionally been applied to associated problems such as the generation and transformation of images and speech signals. These can encode a raw image (comprised of a large number of pixels) into a low-dimensional latent vector. With this example, we explore the potency of our framework, which in essence comprises a dynamical VAE, on vision-based measurements and try to estimate the low-dimensional latent state for disentangling features within the image. In this particular pendulum example, the goal is to infer the angle and angular velocity (latent state $\textbf{z}$) of a dynamic pendulum from a series of images (observations $\textbf{x}$).}

A simple pendulum system is governed by the following differential equation:
\begin{equation}\label{eq:pendulum}
ml^2\ddot{\theta}(t)=-\mu\dot{\theta}(t)-mgl\sin\theta(t),
\end{equation}
where $m=1, l=1,\mu=0.5,g=9.8$. We simulate the system's response, in terms of the angular displacement $\theta(t)$, by applying numerical integration (Runge-Kutta method) with a time step $\Delta t = 0.1$, We subsequently generate image observations from the ground-truth angle $\theta(t)$ to simulate observations from vision-based systems. The images are transformed in black-and-white format and subsequently down-sampled into $16\times 16$ pixels, thus forming observations of binary values of dimension $\mathbb{R}^{n_x}=256$. The mapping from $\theta(t)$ to the image-based observation is non-trivial to formulate as a closed-form function. 
\subsection*{Physics-guided model}
In this example, we adopt the linearization of the pendulum Eq.\eqref{eq:pendulum} as an approximative model for partially capturing the underlying physics:

\begin{equation}\label{eq:pendulum_linear}
ml^2\ddot{\theta}(t)=-\mu\dot{\theta}(t)-mgl\theta(t),
\end{equation}
This can be further cast into a continuous state-space form with $\textbf{z} = [\theta, \; \dot{\theta}]^T$:
$
    \dot{\textbf{z}} = \textbf{A}_c\textbf{z}
$
, where $\textbf{A}_\textbf{c}=\begin{bmatrix} 0 & 1 \\
-9.8 & -0.5
\end{bmatrix}
$. The discretized state-space equivalent model is then obtained as:
\begin{equation}
f_\textbf{phy}: \;\; \textbf{z}_{t}^{\textbf{phy}} = \textbf{A}\textbf{z}_{t-1}^\textbf{phy}.
\end{equation}
with the discrete matrix standardly computed as $\textbf{A}=\textbf{e}^{\textbf{A}_\textbf{c}\Delta t}$, thus delivering the physics-guided model term used for this example. It is noted that in Eq.\eqref{eq:pendulum_linear}, a small angle approximation has been made, implying that $\sin\theta\approx\theta$. This may not hold for larger angles, however, the benefit of the suggested approach is that the approximation need not be highly accurate, as the deep learning-based model term is expected to capture the resulting discrepancy. 

\subsection*{Emission model}
The emission process is here modelled as an independent multivariate Bernoulli distribution, since the observations are binary images, with the parameters of the Bernoulli distribution learned via the neural network function $\mathcal{NN}_2$ as illustrated in Section \ref{sec:emission}, i.e.,
\begin{equation}
\textbf{x}_t\sim \text{Bernoulli}(\mathcal{NN}_2(\textbf{z}_t)).
\end{equation}
\subsection*{Results}
\begin{figure}[htbp]
\centering
\begin{subfigure}{.5\textwidth}
  \centering
  \includegraphics[height=1.0\linewidth]{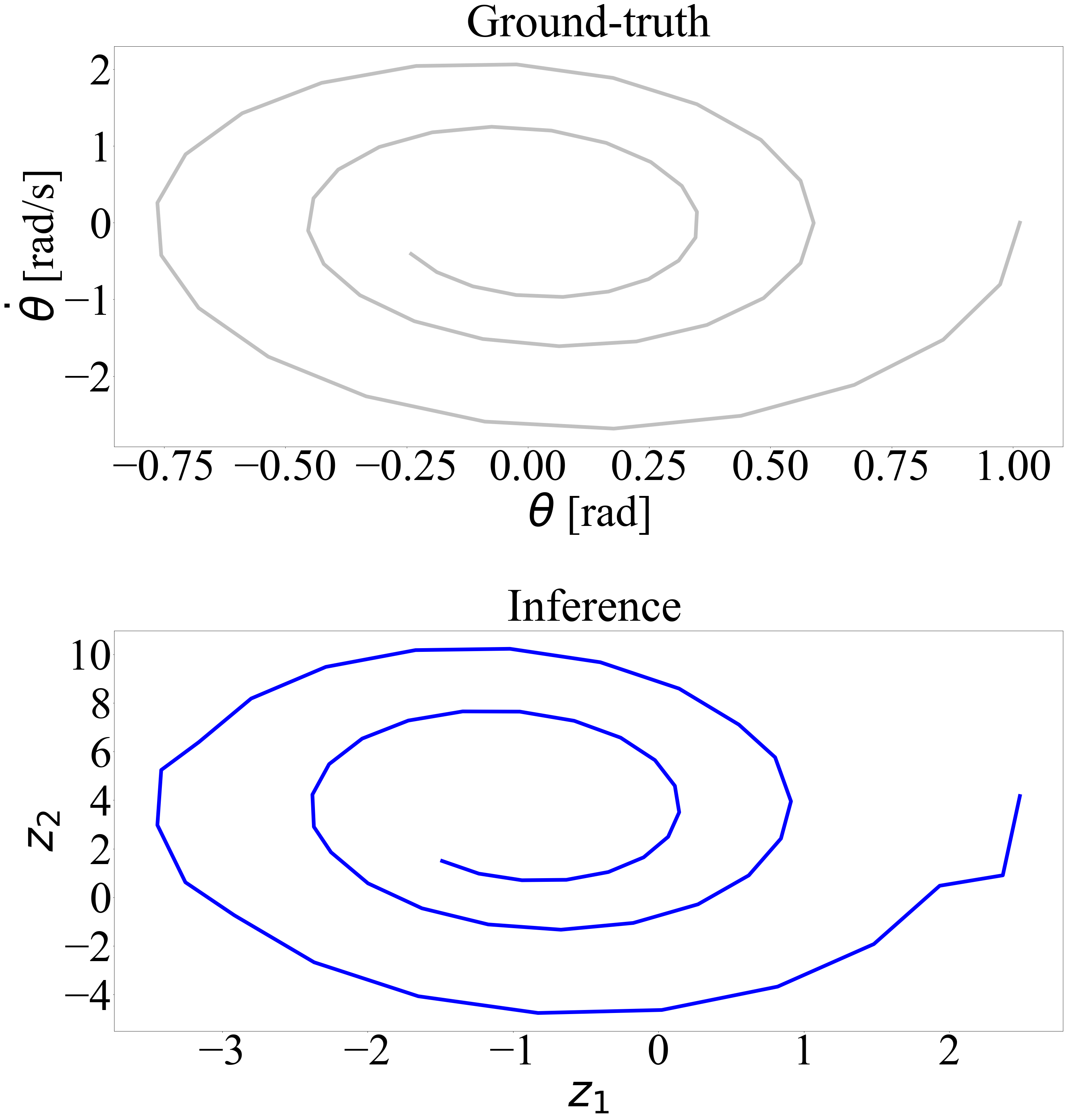}
  \caption{PgDMM}
  \label{fig:pendulum_phase}
\end{subfigure}%
\begin{subfigure}{.5\textwidth}
  \centering
  \includegraphics[height=1.0\linewidth]{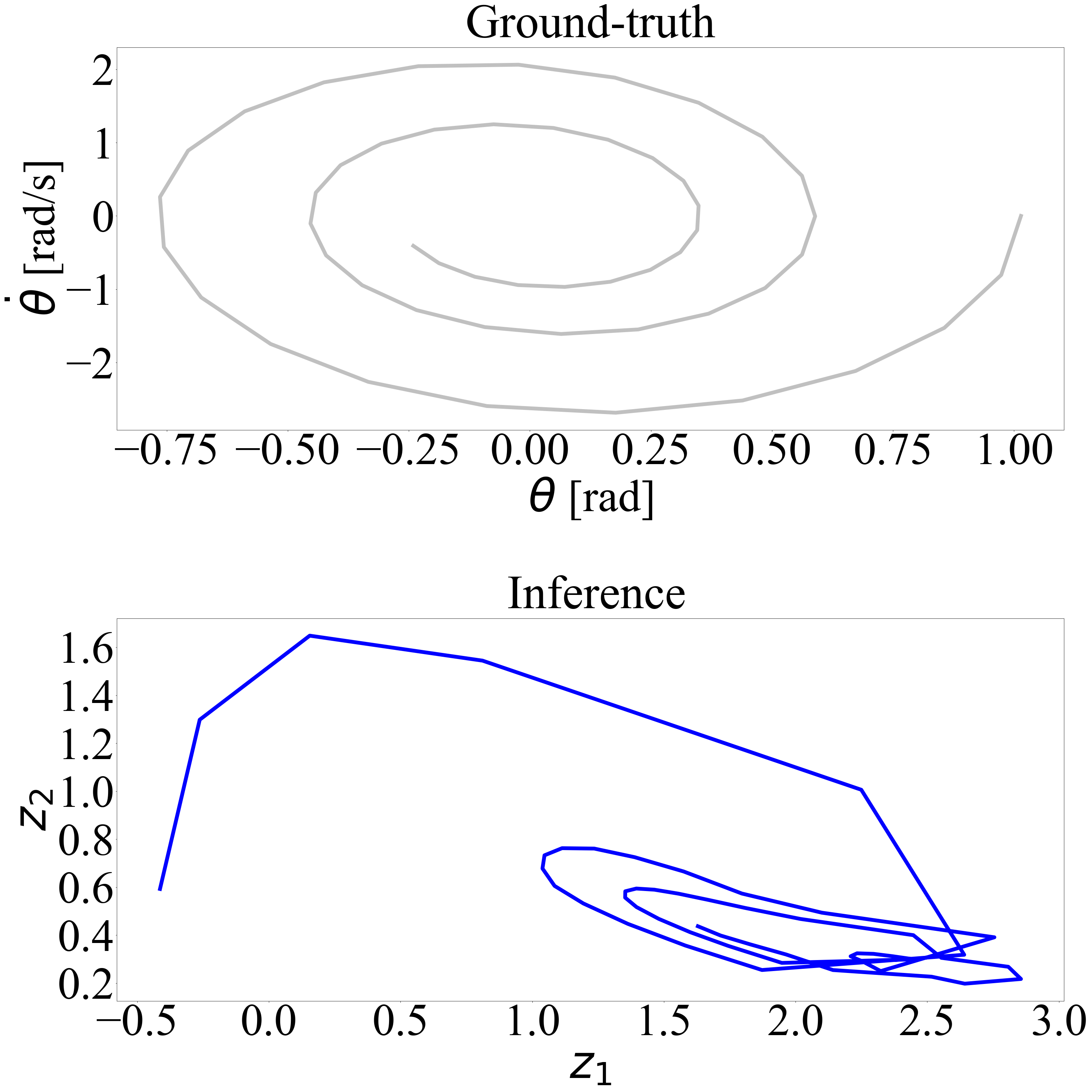}
  \caption{DMM}
  \label{fig:pendulum_phase_dmm}
\end{subfigure}
\begin{subfigure}{1.0\textwidth}
  \centering
  \includegraphics[height=0.1\linewidth]{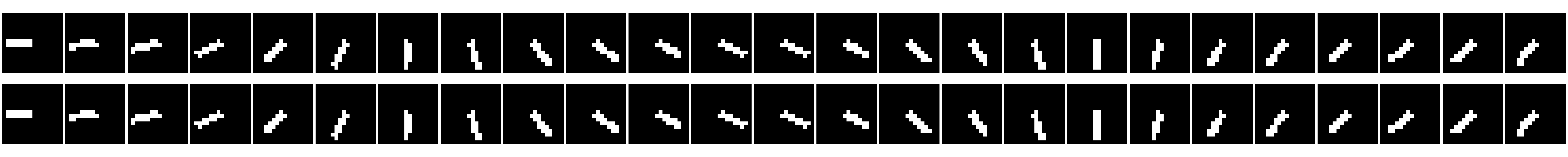}
  \caption{}
  \label{fig:pendulum_images}
\end{subfigure}
\caption{Training results of the pendulum example: (a) phase portrait of ground-truth (top) and PgDMM (bottom); (b) phase portrait of ground-truth (top, repeated) and DMM (bottom). The results for the DMM are obtained by fixing the weight $\alpha=0$, i.e., only a learning-based model; (c) ground-truth observations images (top) and reconstructed images (bottom) from PgDMM. Reconstructed images are obtained by feeding the inferred latent states into the learned emission function.}
\label{fig:pendulum_phase_train}
\end{figure}

Figure \ref{fig:pendulum_phase} compares the phase portraits, i.e., the plot of angular velocity, $\dot{\theta}$, versus displacement, $\theta$, for the pendulum system. The ground truth (top subplot) is compared against the latent states $z_1$ versus $z_2$ (bottom subplot), as inferred by the proposed PgDMM model. {\color{black} The units of the latent states are not added since these are inferred values which are not guaranteed to be physical coordinates, although, as we point out, one of the advantages of the proposed PgDMM lies in rendering them physically interpretable.} It is observed that the learned latent space precisely follows the ground truth, albeit with a different scaling of the latent states and a shift in terms of the center. This can be explained by the fact that the emission process is purely parameterized by a neural network and it is not guaranteed that the trained emission model exactly mimics the process generation we followed for image-based observations. However, a physically disentangled latent space is indeed captured, with the trained emission and transition models thus forming a scaled/shifted version of the true transition and emission processes. Figure \ref{fig:pendulum_images} illustrates the first 25 steps of the reconstructed observation images, i.e., the images generated from the latent states learned by inference model, which highly agrees with the ground truth. On the other hand, as shown in Figure \ref{fig:pendulum_phase_dmm}, the phase portrait of the latent space  which is learned by the DMM (bottom subplot), fails to yield a physically interpretable latent space, as this is not a requirement for this type of learning framework.

\begin{figure}[htbp]
\centering
\begin{subfigure}{.5\textwidth}
  \centering
  \includegraphics[height=0.5\linewidth]{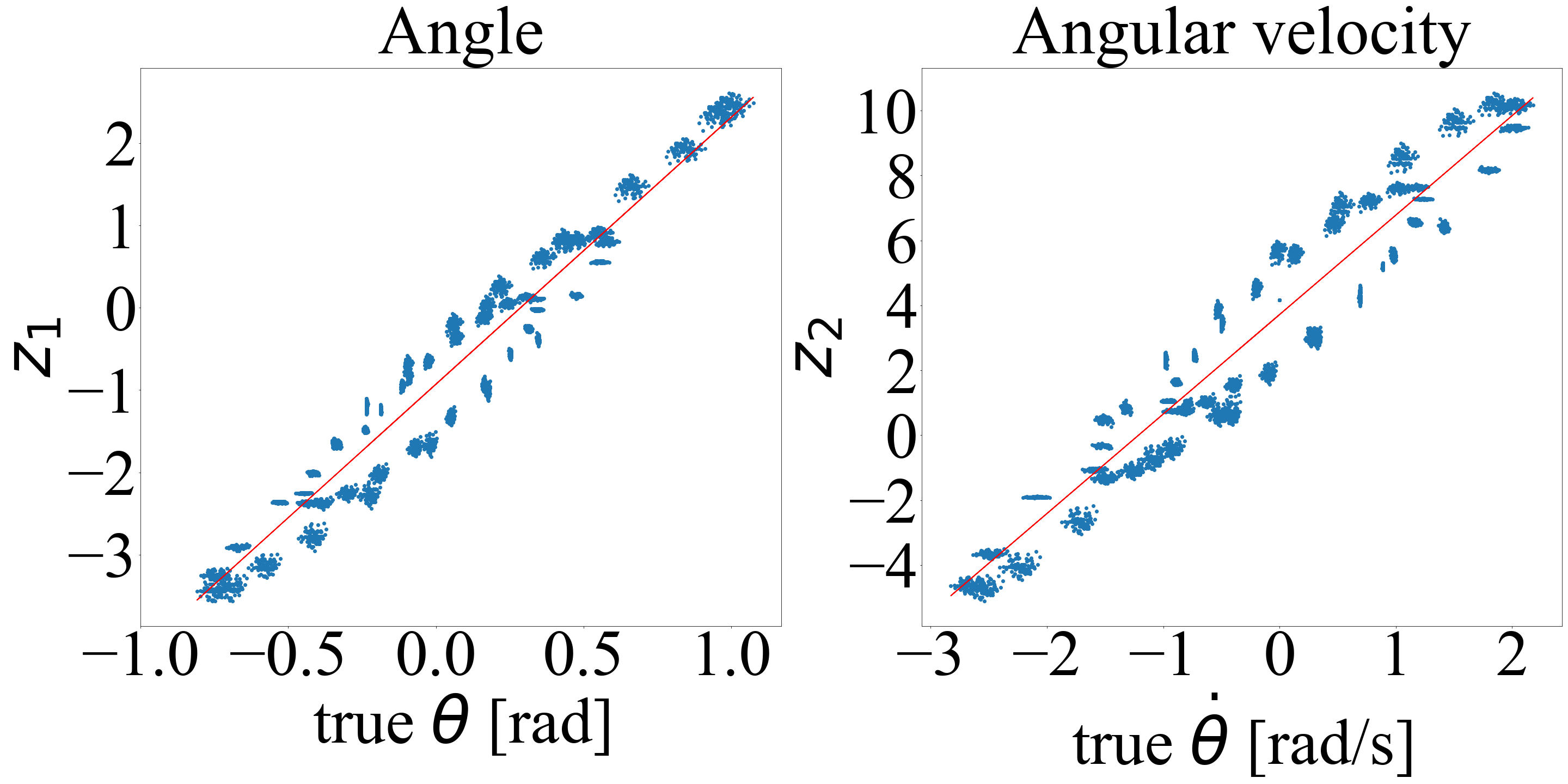}
  \caption{PgDMM}
  \label{fig:pendulum_reg}
\end{subfigure}%
\begin{subfigure}{.5\textwidth}
  \centering
  \includegraphics[height=0.5\linewidth]{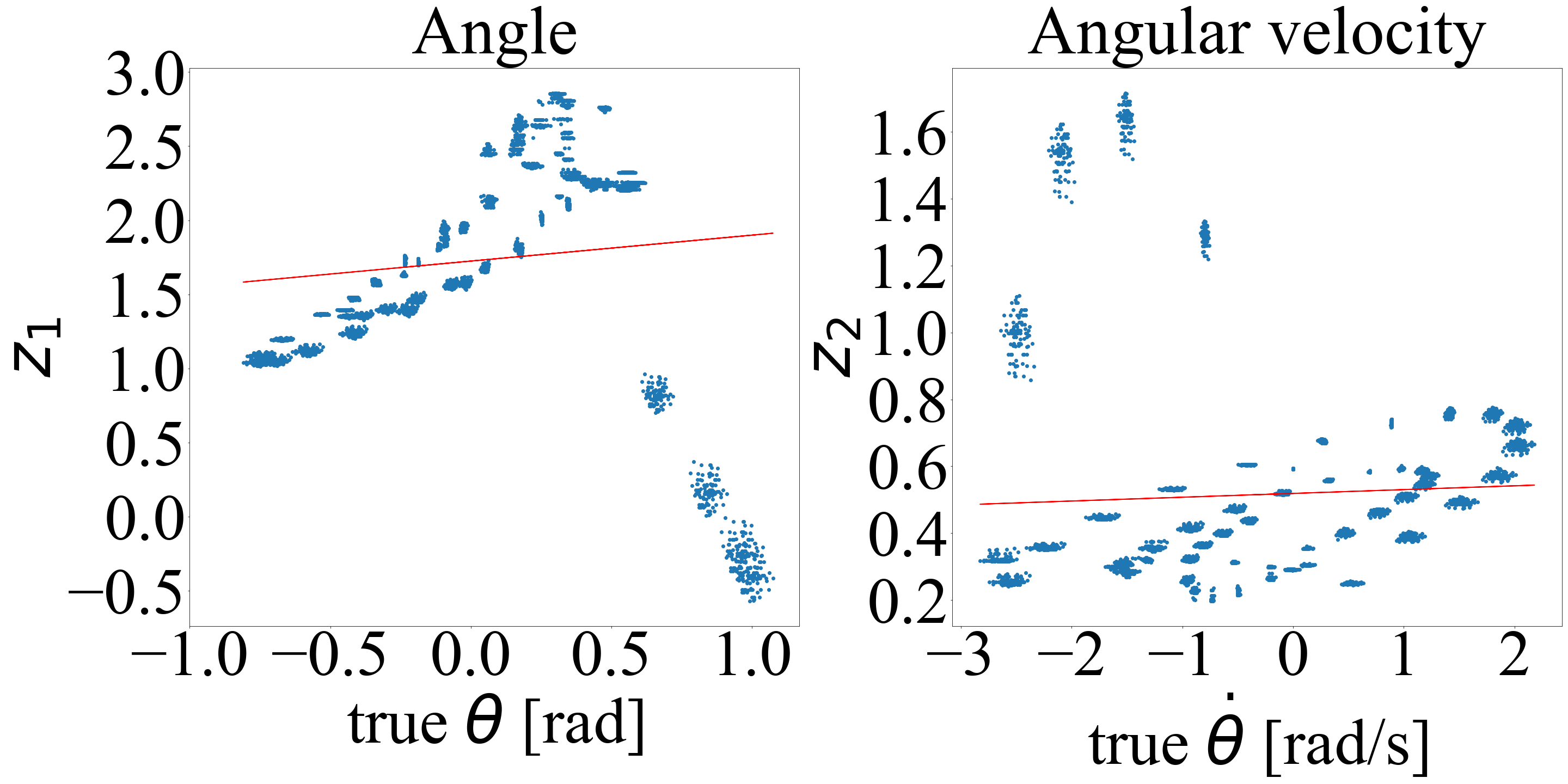}
  \caption{DMM}
  \label{fig:pendulum_reg_dmm}
\end{subfigure}
\caption{Regression results with training data (pendulum example). {\color{black} The fitted linear regression lines are plotted in red. (a) The scatter plot and regression line of the ground-truth versus the latent states learned from the PgDMM model. (b) The scatter plot and regression line of the ground-truth versus the latent states learned from the DMM model.}}
\label{fig:pendulum_reg_train}
\end{figure}

The regression analysis of the inferred versus true states, based on an ordinary least squares (OLS) fit, for both the angle and the angle velocity variables are shown in Figure \ref{fig:pendulum_reg_train}. The OLS regression results further validate what is observed in Figure \ref{fig:pendulum_phase_train}, namely a strong correlation between learned latent states and the ground-truth angle and angular velocity variables. Both goodness-of-fit scores, $R^2$, lie very close to $1$, indicating the superior performance of the PgDMM. The DMM features a lower score, as no boosting of a-prior knowledge of the physics is here exploited for gearing the model structure toward an explainable latent space.

\begin{figure}[htbp]
\centering
\begin{subfigure}{.5\textwidth}
  \centering
  \includegraphics[height=1.0\linewidth]{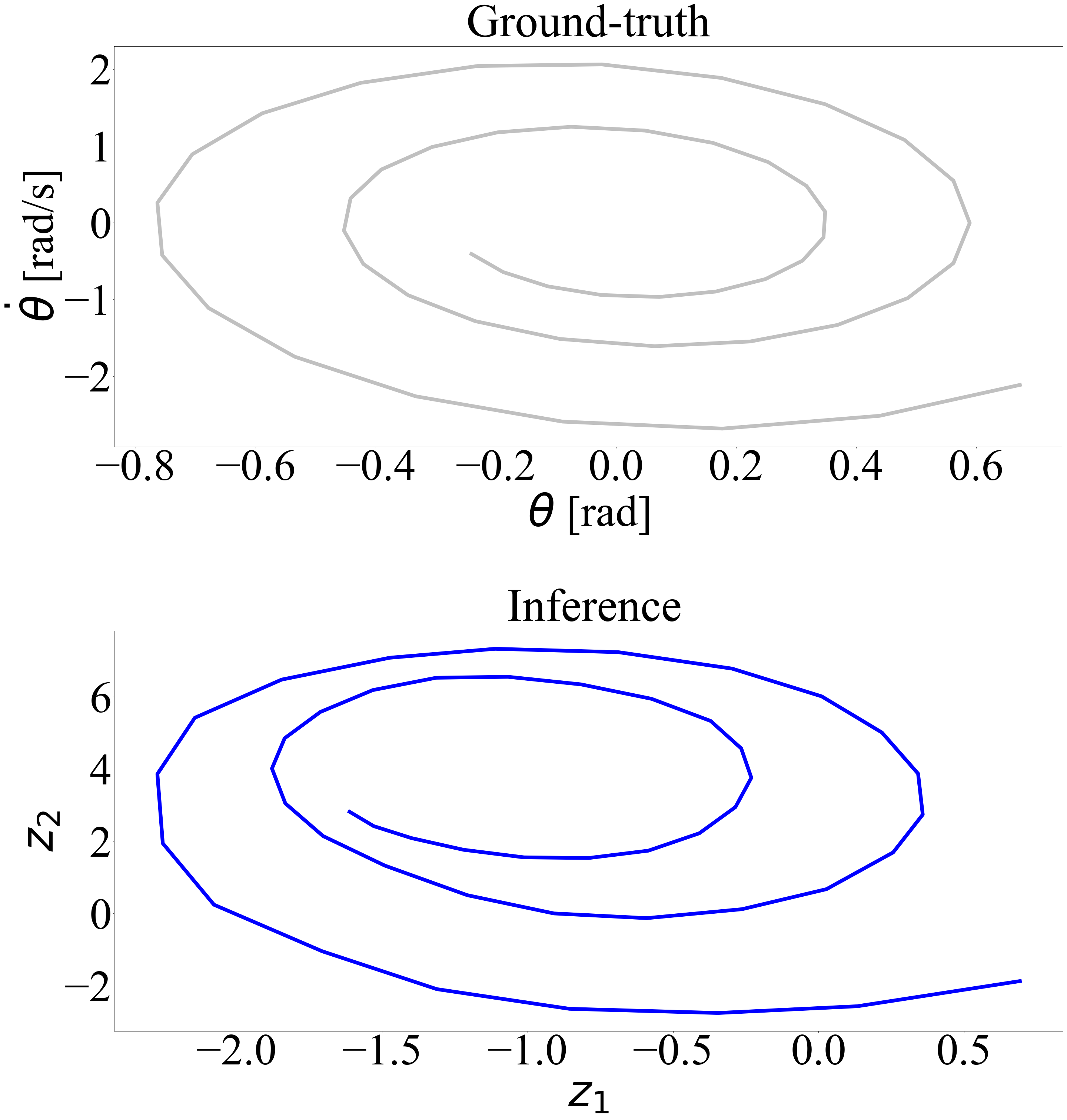}
  \caption{PgDMM}
  \label{fig:pendulum_phase_test}
\end{subfigure}%
\begin{subfigure}{.5\textwidth}
  \centering
  \includegraphics[height=1.0\linewidth]{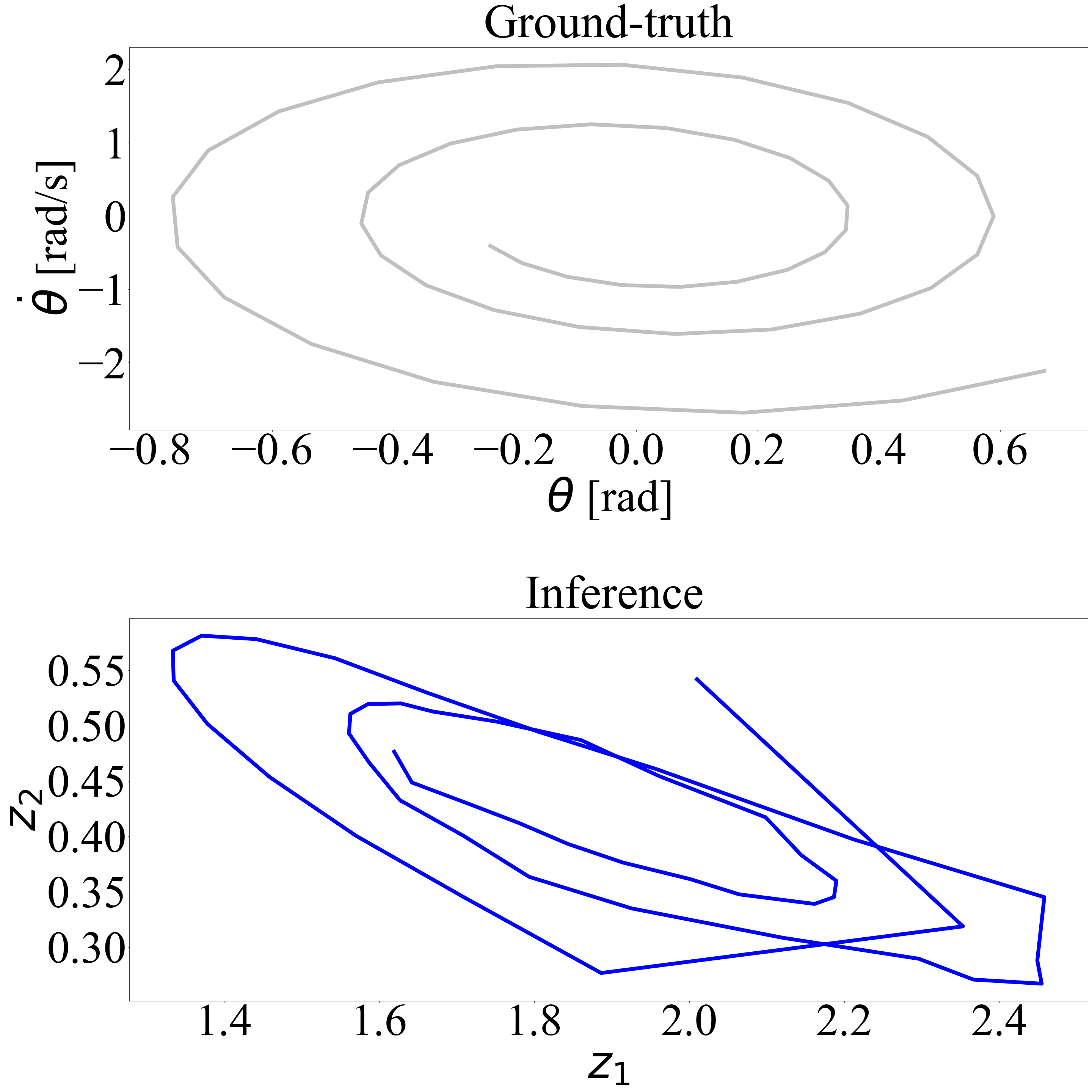}
  \caption{DMM}
  \label{fig:pendulum_phase_dmm_test}
\end{subfigure}
\begin{subfigure}{1.0\textwidth}
  \centering
  \includegraphics[height=0.1\linewidth]{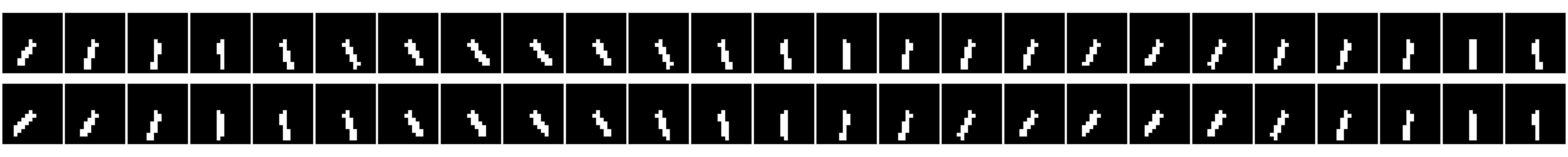}
  \caption{}
  \label{fig:pendulum_images_test}
\end{subfigure}
\caption{Testing results of the pendulum example: (a) phase plot of ground-truth (top) and PgDMM (bottom) on the test dataset. This result on the test dataset shows the generalization and prediction capabilities for data beyond the scale of training sequences; (b) phase plot of ground-truth (top) and DMM (bottom) on the test dataset; (c) ground-truth observations images (top) and reconstructed images from PgDMM (bottom) on the test dataset.}
\label{fig:pendulum_test_phase}
\end{figure}

\begin{figure}[htbp]
\centering
\begin{subfigure}{.5\textwidth}
  \centering
  \includegraphics[height=0.5\linewidth]{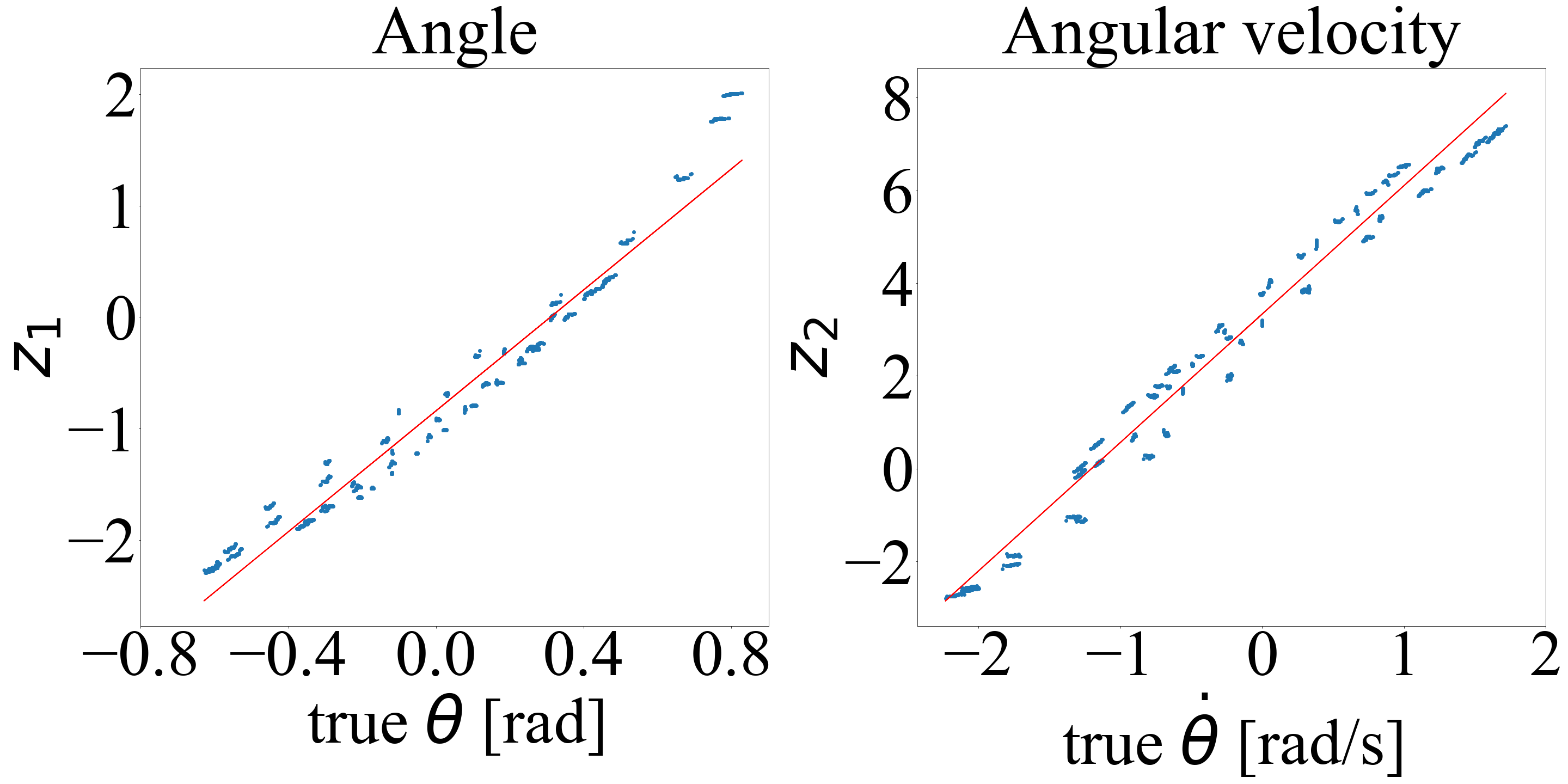}
  \caption{PgDMM}
  \label{fig:pendulum_reg_test}
\end{subfigure}%
\begin{subfigure}{.5\textwidth}
  \centering
  \includegraphics[height=0.5\linewidth]{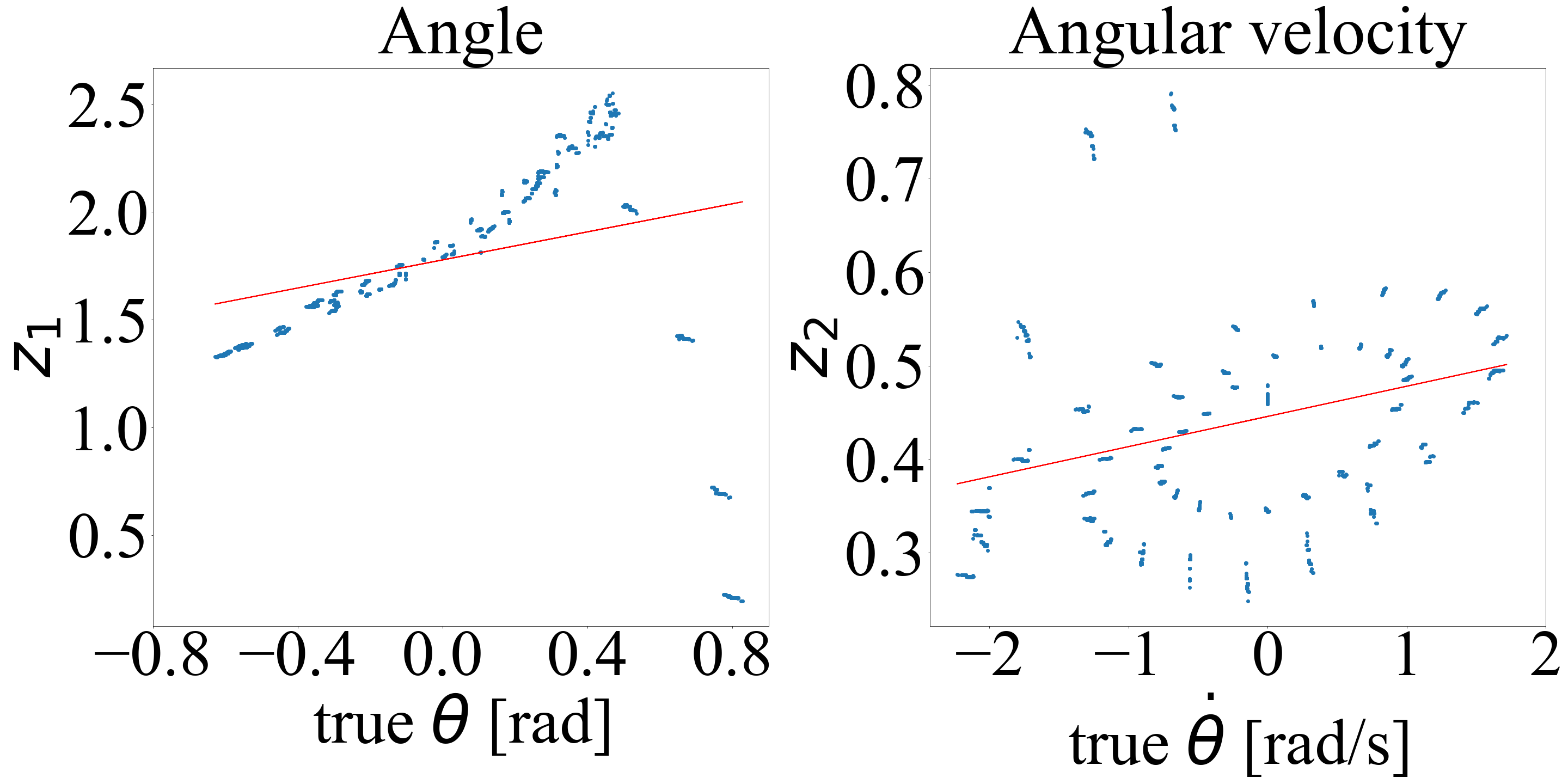}
  \caption{DMM}
  \label{fig:pendulum_reg_dmm_test}
\end{subfigure}
\caption{Regression results with test data (pendulum example). {\color{black} The fitted linear regression lines are plotted in red. (a) The scatter plot and regression line of the ground-truth and the latent states learned from the PgDMM model. (b) The scatter plot and regression line of the ground-truth and the latent states learned from the DMM model.}}
\label{fig:pendulum_test_reg}
\end{figure}

We lastly generated a further set of image sequences, instantiated from different initial conditions, to serve as test data. The performance is similar for the validation on the test dataset, as shown in Figure \ref{fig:pendulum_test_phase}. A strong correlation, as shown in Figure \ref{fig:pendulum_test_reg}, is obtained between the learned latent states and ground-truth, also in testing mode. This indicates that the PgDMM trained model indeed comprises a physically interpretable latent space, which further proves beneficial for prediction, especially when compared against the DMM performance.

\subsection{Fatigue Crack Growth}
As a second case study, We simulate a progressive crack growth problem, using the following analytical equations \cite{corbetta2018optimization} based on Paris' law:
\begin{equation}\label{eq:crack}
\begin{split}
z_t&=z_{t-1}+C(\Delta S\sqrt{z_{t-1}\pi})^m\Delta N\cdot e^{\omega_t}, \\
x_t&=z_t+\nu_t,
\end{split}
\end{equation}
where Paris' law parameters $C=e^{-33}$, $m=4$, applied fatigue stress $\Delta S=60$, and load cycles for one step $\Delta N=1400$ \cite{corbetta2018optimization}. The one-dimensional latent variable $z_t$ represents the true crack length at time $t$, with the model errors (or process noise) modelled as a log-normal random process $e^{\omega_t}$, where $\omega_t\sim\mathcal{N}(0,0.1)$, multiplied by a constant rate $\Delta N$. The observation $x_t$ represents the measurement of the crack length at time $t$ with an assumed measurement noise $\nu_t \sim \mathcal{N}(0, 0.04)$. If we consider $e^{\omega_t}\approx 1+\omega_t$, Eq.\eqref{eq:crack} can be approximated as:
\begin{equation}\label{FCG}
z_t = z_{t-1}+C(\Delta S\sqrt{z_{t-1}\pi})^m\Delta N+C(\Delta S\sqrt{z_{t-1}\pi})^m\Delta N\omega_t.
\end{equation}
Based on Eq.\eqref{FCG}, in this example, the physics-guided model is given by:
\begin{equation}
f_\textbf{phy}: \; z_{t} = z_{t-1}+C(\Delta S\sqrt{z_{t-1}\pi})^m\Delta N.
\end{equation}
We simulate $200$ realizations of observation sequences $x_{1:T}$ with length $T=100$ using Eq.\eqref{FCG}. The first $60$ time steps serve for training, while the last $40$ time steps are retained for testing. It noted that, due to the nature of the process, the amplitude of the test dataset is higher than the amplitude of the training dataset.
\begin{figure}
\centering
\begin{subfigure}{.5\textwidth}
  \centering
  \includegraphics[height=0.7\linewidth]{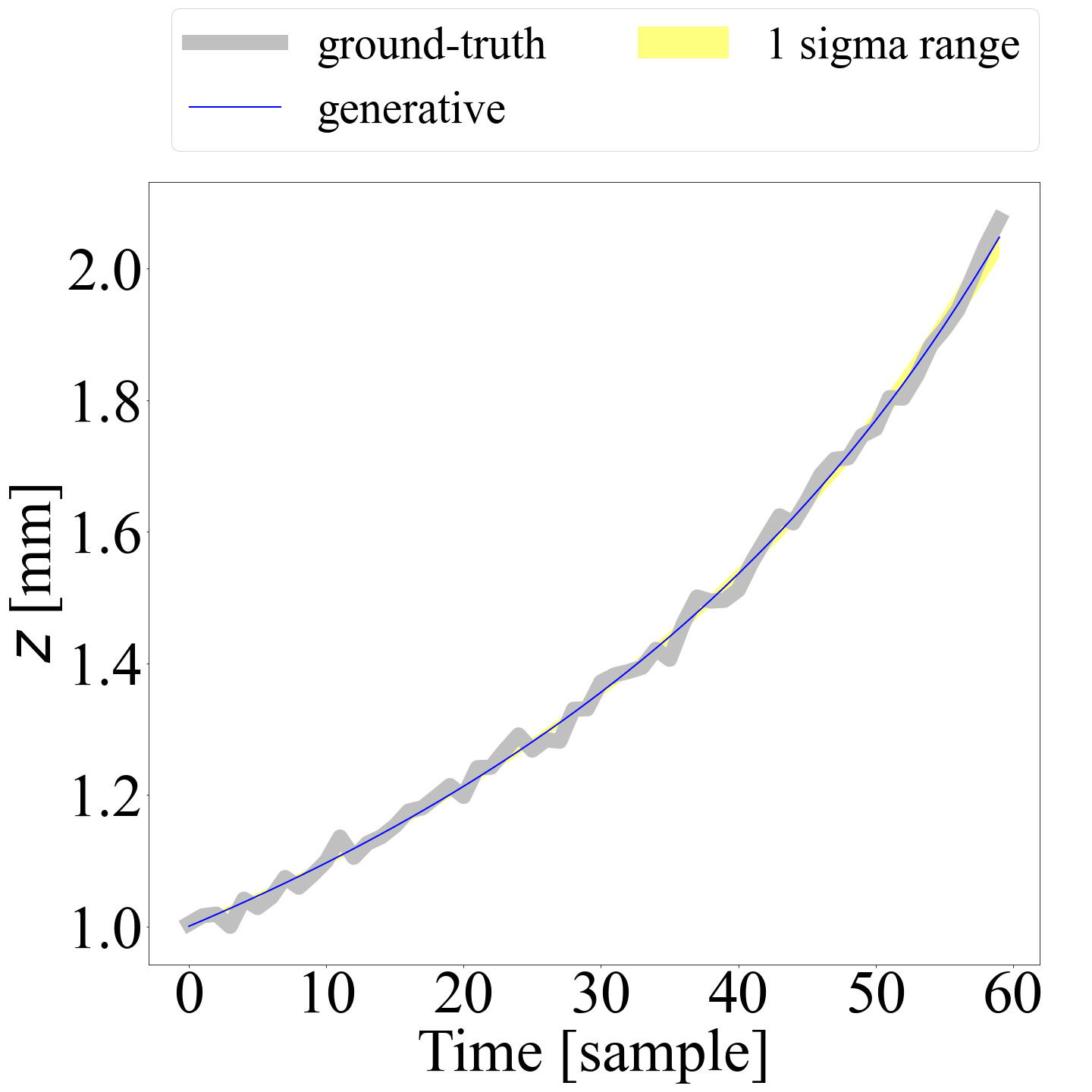}
  \caption{Inferred latent states\\\hspace{\textwidth}}
  \label{fig:crack_gen}
\end{subfigure}%
\begin{subfigure}{.5\textwidth}
  \centering
  \includegraphics[height=0.7\linewidth]{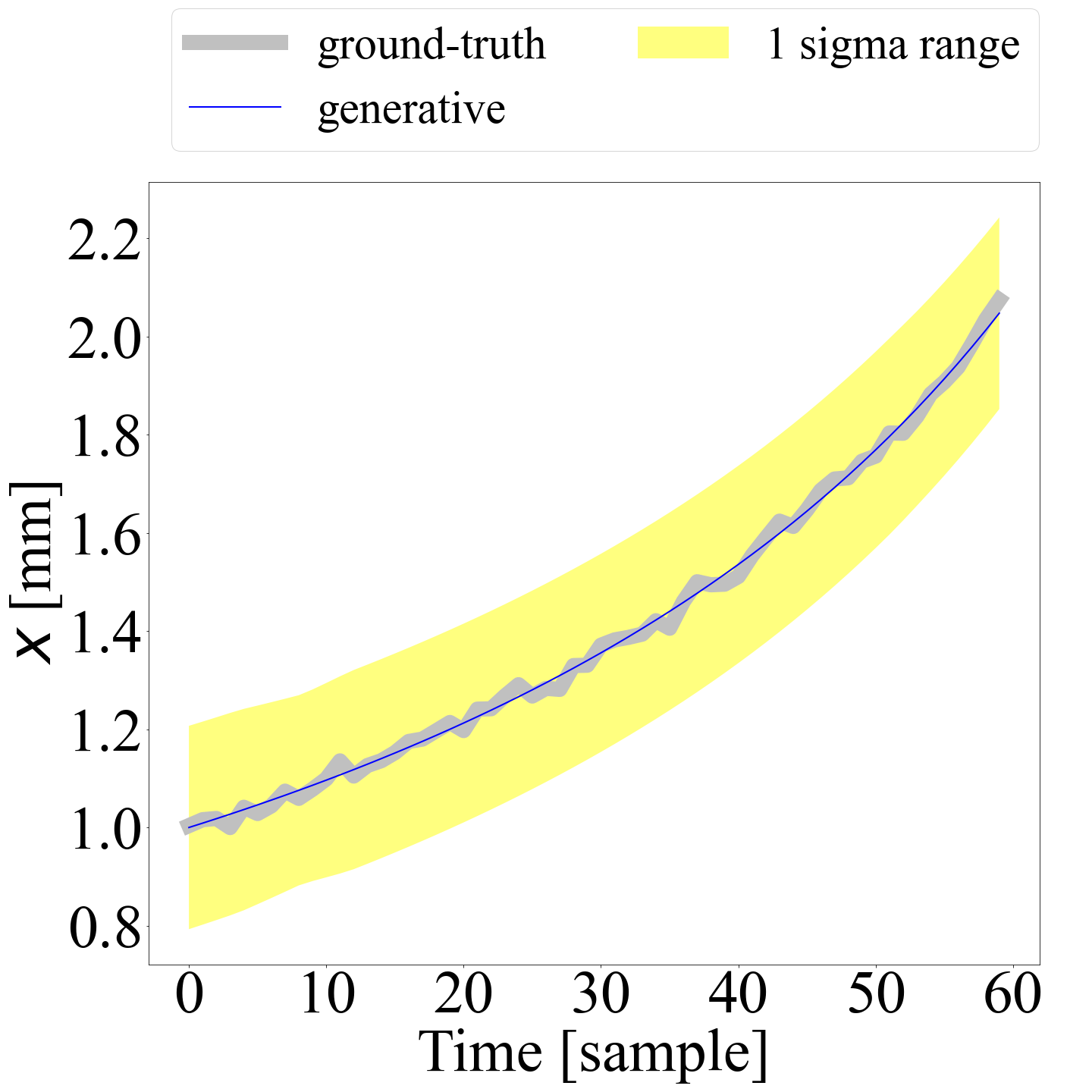}
  \caption{Reconstructed observations\\\hspace{\textwidth}}
  \label{fig:crack_obs}
\end{subfigure}
\vfill
\begin{subfigure}{.5\textwidth}
  \centering
  \includegraphics[height=0.7\linewidth]{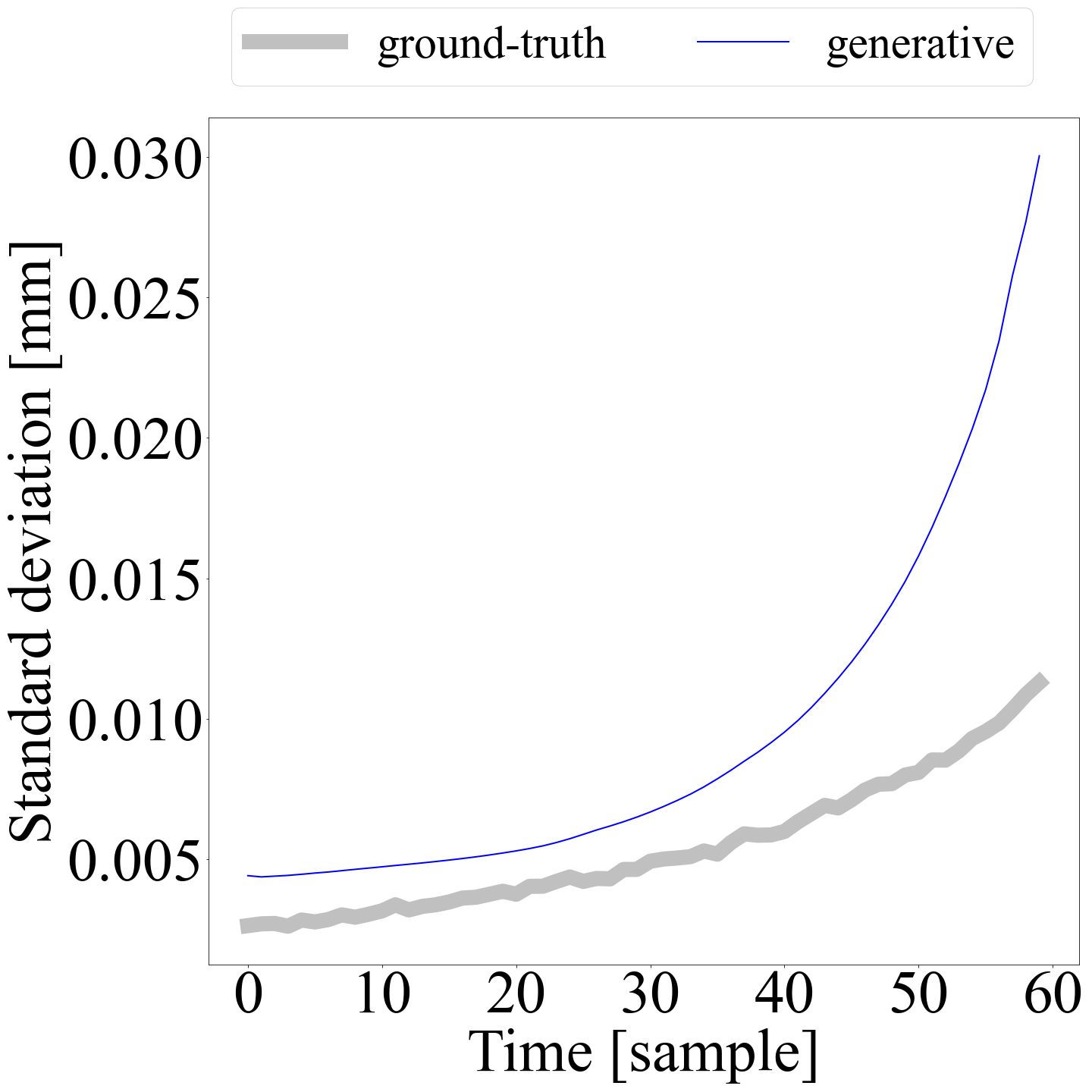}
  \caption{Inferred standard deviations of transition process noises}
  \label{fig:crack_var}
\end{subfigure}%
\begin{subfigure}{.5\textwidth}
  \centering
  \includegraphics[height=0.7\linewidth]{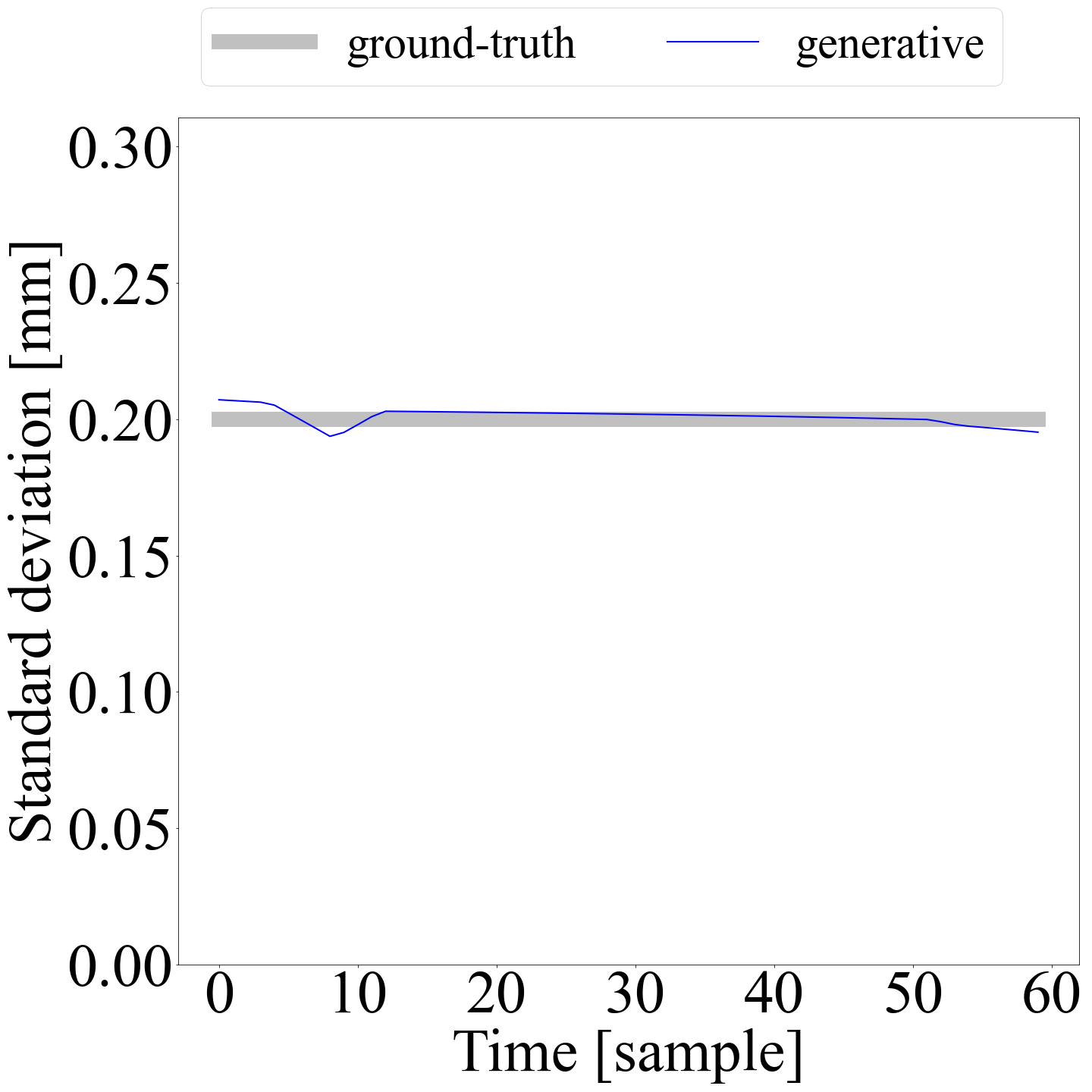}
  \caption{Inferred standard deviations of emission process noises}
  \label{fig:crack_obs_var}
\end{subfigure}
\caption{Training results for the crack growth problem. The estimation results are compared against the corresponding ground-truth. The yellow shaded area in Figures \ref{fig:crack_gen} and \ref{fig:crack_obs} denotes the inferred standard deviations of the transition and emission process noises respectively, as further plotted in Figure \ref{fig:crack_var} and \ref{fig:crack_obs_var}.}
\label{fig:crack_latent}
\end{figure}
In Figures \ref{fig:crack_gen} and \ref{fig:crack_obs}, we show the training results of a realization in terms of the mean values of the learned latent space ($z_t$) and the reconstructed observation ($x_t$), respectively. Compared to the ground-truth, it is observed that the learned latent space and reconstructed observation capture the trend of the progressive crack growth. 

We further investigate the capability of the proposed PgDMM in capturing the process uncertainty embedded in Eq.\eqref{FCG}. Note that the transition process uncertainty features an explicit distribution given as:
\begin{subequations}
    \begin{equation}
C(\Delta S\sqrt{z_{t-1}\pi})^m\Delta N\omega_t\sim\mathcal{N}(0, \sigma^2),
\end{equation}
in which,
\begin{equation}\label{eq:sigma}
    \sigma = C(\Delta S\sqrt{z_{t-1}\pi})^{m}\Delta N\cdot \sqrt{0.1},
\end{equation}
\end{subequations}
where the standard deviation $\sigma$ of the uncertainty is a function of $z_{t-1}$, reflecting an increasing variance with time. Since the fatigue crack growth problem is one-dimensional, the standard deviation learned by PgDMM is computed as the square root of the covariance matrix given in Eq.\eqref{mean_var}. The standard deviations learned by the model are compared to the ground-truth (computed by Eq.\eqref{eq:sigma}) in Figure \ref{fig:crack_var}, where the inferred standard deviations reliably follow the increasing trend of the ground truth. A similar result for emission process uncertainty is shown in Figure \ref{fig:crack_obs_var}. Note that the standard deviation of the emission process noise $\nu_t$ is simply $0.2$ and it can be observed that there is a strong agreement between the inferred standard deviations and the ground-truth for the emission process uncertainty.

\begin{figure}[h]
  \centering
  \includegraphics[height=0.35\linewidth]{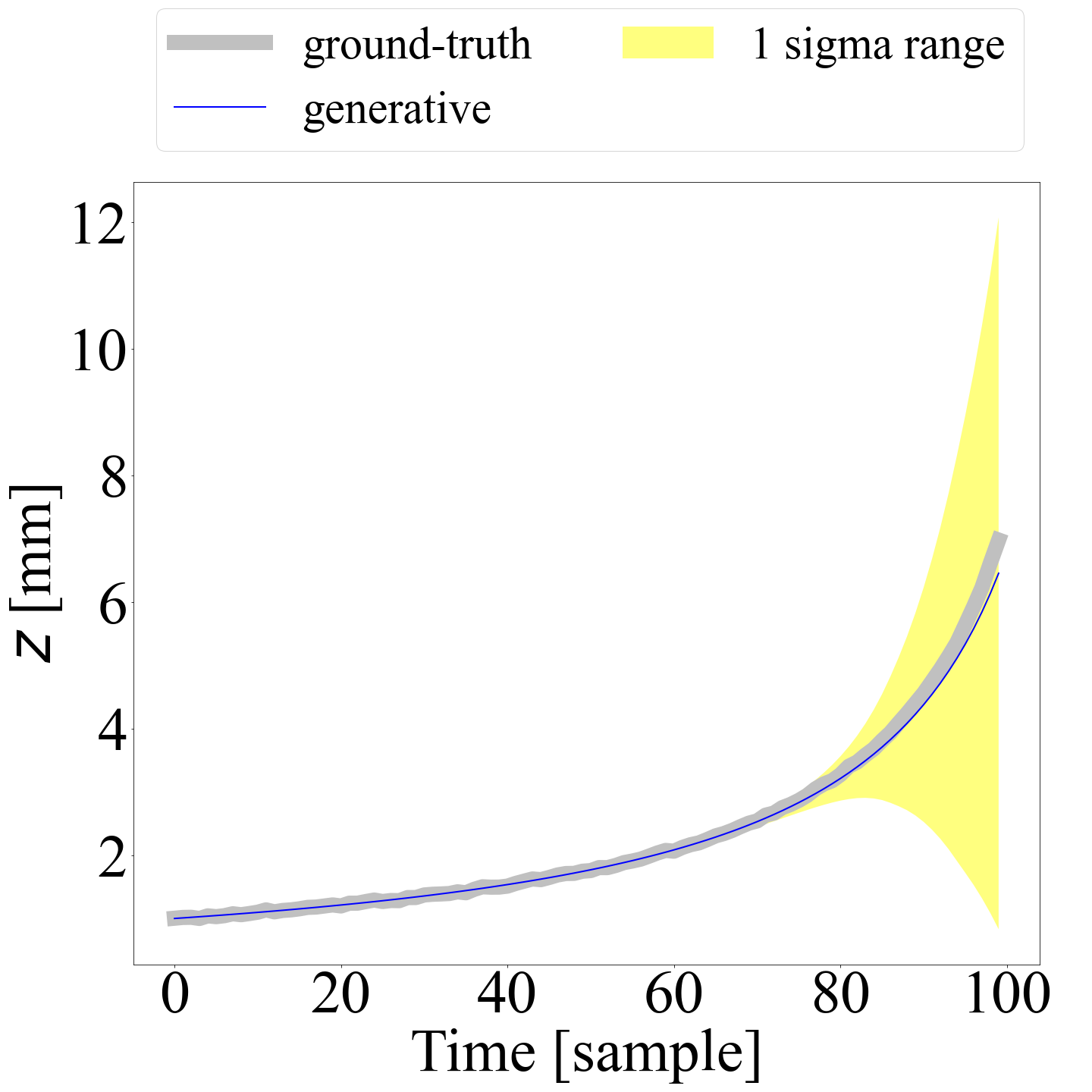}
\caption{Inferred latent states with test data. Yellow shaded area denotes standard deviations of the inference model.}
  \label{fig:crack_gen_test}
\end{figure}

Furthermore, the testing result is shown in Figure \ref{fig:crack_gen_test}. The complete sequence is shown, where the first $60$ time steps are the training result, same as Figure \ref{fig:crack_gen}, and the last $40$ time steps are the testing result. We can observe that the learned latent states fit the ground-truth well even beyond the training sequence, albeit with much larger uncertainty, revealing that the model also works for prediction in this example.

\subsection{Experiment Result: Silverbox System Benchmark}
As a final example we here adopt the Silverbox benchmark problem \cite{wigren2013three}. This is an electrical circuit that emulates the behavior of a nonlinear mass-spring-damper system and the governing equation is approximately given by the following second-order differential equation \cite{wigren2013three}:
\begin{align}
m\ddot{x}(t)+c\dot{x}(t)+kx(t)+k_nx^3(t)=u(t),
\end{align}
which is close to the behavior of a Duffing oscillator. In this benchmark dataset a known force $u(t)$ is applied to the system and the resulting displacement $x(t)$ is measured; $k$ and $k_n$ describe the linear and nonlinear spring stiffness parameters, while $c$ denotes the viscous damping coefficient. 
\begin{figure}[h]
	\centering
    \includegraphics[width=0.75\linewidth]{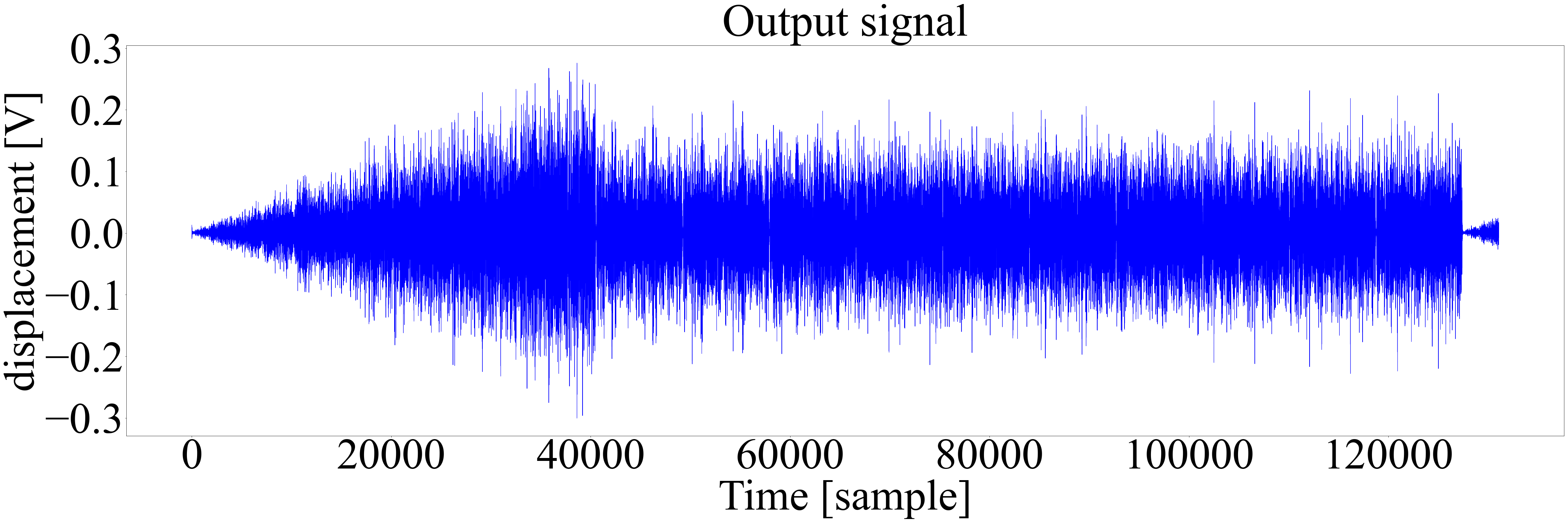}
	\caption{The dataset of the Silverbox benchmark problem. The data from $t=40,001$ to $t=100,000$, the tail of the arrow, are used for training, while data from $t=1$ to $t=40,000$, the head of the arrow, are retained for testing. {\color{black} The Silverbox benchmark problem is an electrical circuit that emulates the behavior of a nonlinear mass-spring-damper system; the units of the ``displacement-like" quantities are in Volts.}}
	\label{fig:silverbox_data}
\end{figure}

The full-length output data are shown in Figure \ref{fig:silverbox_data}, forming an arrow shape. We use the data from $t = 40,001$ to $t = 100,000$ (the tail shape part) for training, while data from $t = 1$ to $t = 40,000$ (the head shape part) are retained for testing. For training the model using a mini-batch strategy, the training data are subdivided into 400 mini-batches,  with uniform length equal to 100.

We first estimate an approximate linear model as the physics-guided model term to infuse into this system. The parameters are obtained via the following differential equation:
\begin{equation}
m\ddot{x}(t)+c\dot{x}(t)+kx(t) = u(t),
\end{equation}
{\color{black} where the estimated $m=5\times10^{-6}$, $c=2.4892\times10^{-4}$, $k=0.9436$.} This can be further cast into a continuous state-space form with $\textbf{z}(t) = [x(t), \; \dot{x}(t)]^T$:
\begin{equation}
\begin{split}
\dot{\textbf{z}}(t) &= \textbf{A}_c\textbf{z}(t)+\textbf{B}_cu(t),\\
\textbf{x}(t) &= \textbf{C}\textbf{z}(t),
\end{split}
\end{equation}
where 
$\textbf{C}=[1, \; 0]$.
This continuous state-space model is further discretized as
\begin{equation}
\begin{split}
\textbf{z}_t&=\textbf{A}\textbf{z}_{t-1}+\textbf{B}u_t,\\
\textbf{x}_t&=\textbf{C}\textbf{z}_t.
\end{split}
\end{equation}
We make use of the matrix \textbf{C} for the emission process and matrices \textbf{A} and \textbf{B} for formulating the transition process of the physics-guided stream of our PgDMM model setup, thus, assuming a suspected approximation of the actual dynamics, as follows:
\begin{equation}\label{eq:silver_phy}
f_\textbf{phy}: \; \textbf{z}_t = \textbf{A}\textbf{z}_{t-1}+\textbf{B}u_t.
\end{equation}
Due to the included neural network-based discrepancy term, we do not require the physics-based approximation to be highly accurate. We also incorporate the inputs and the corresponding matrix \textbf{B} into the inference network of the physics-based stream, with $\mathbf{\mu}_\phi^\textbf{phy}$ in Eq.\eqref{nn_phi} now computed as:
\begin{equation}
\mathbf{\mu}_{\phi}^\textbf{phy}(\textbf{z}^\textbf{phy}_{t-1},\textbf{x}_{1:T}, u_t)= \mathcal{NN}_\phi^\textbf{phy}(\textbf{h}_{t}^\textbf{phy}) + \textbf{B}u_t.
\end{equation}
Moreover, $u_t$ is incorporated into the learning model, with $\textbf{h}_t^\text{NN}$ formulated as:
\begin{equation}
\textbf{h}_t^\text{NN} = \frac{1}{3}[\textbf{h}_t^b+\textbf{h}_t^f + \tanh(\textbf{W}^\text{NN}\begin{bmatrix}\textbf{z}_{t-1}^\text{NN} \\u_t\end{bmatrix}+\textbf{b}^\text{NN})].
\end{equation}
Since the observation dataset only contains the measured displacement $x_t$, we regard it as the ground-truth displacement and compute the difference of displacements to generate velocity data as reference for comparison against the latent space learned from our model.
\begin{figure}[htpb]
\centering
\begin{subfigure}{1\textwidth}
  \centering
  \includegraphics[height=0.56\linewidth]{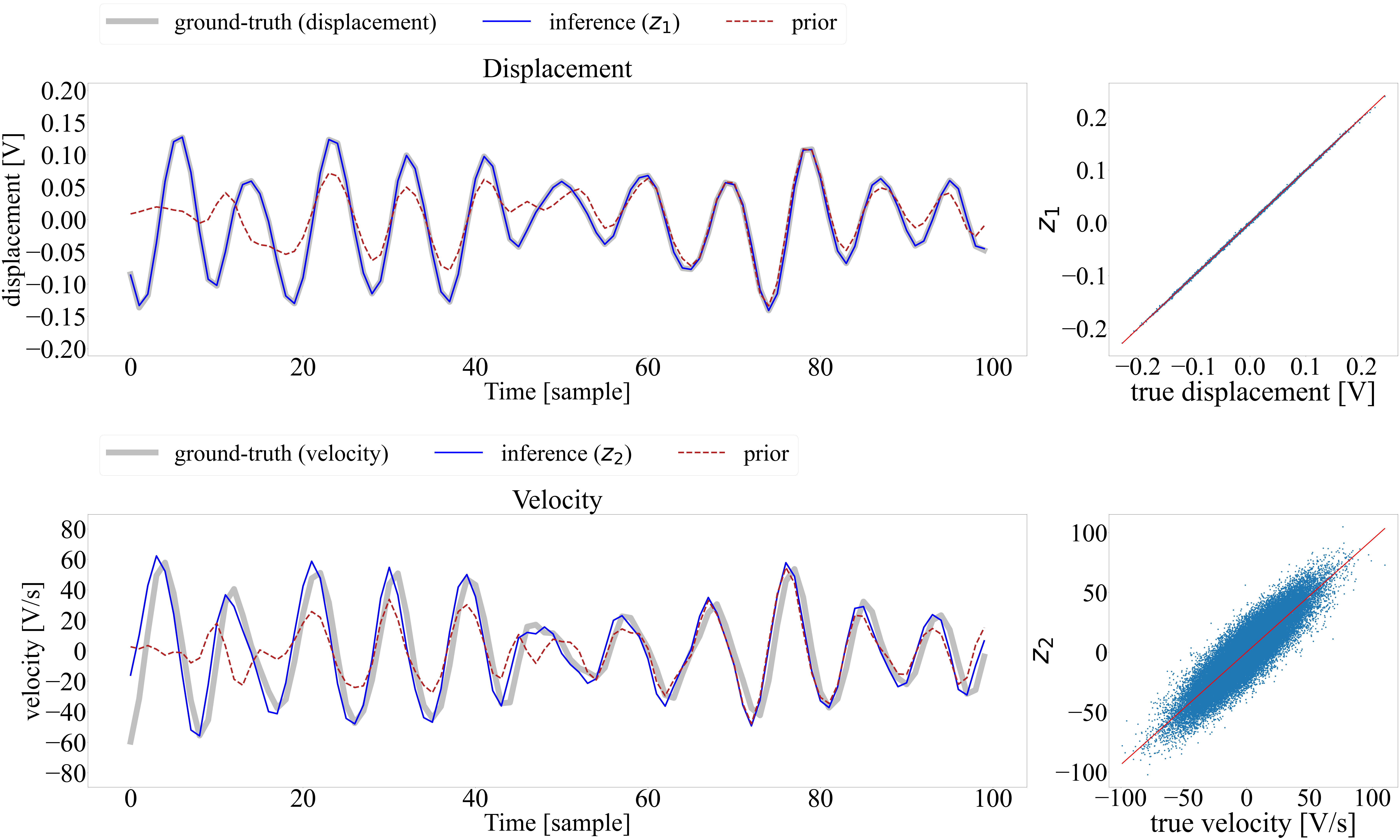}
  \caption{PgDMM}
  \label{fig:silverbox_expanded}
\end{subfigure}
\vfill
\begin{subfigure}{1\textwidth}
  \centering
  \includegraphics[height=0.56\linewidth]{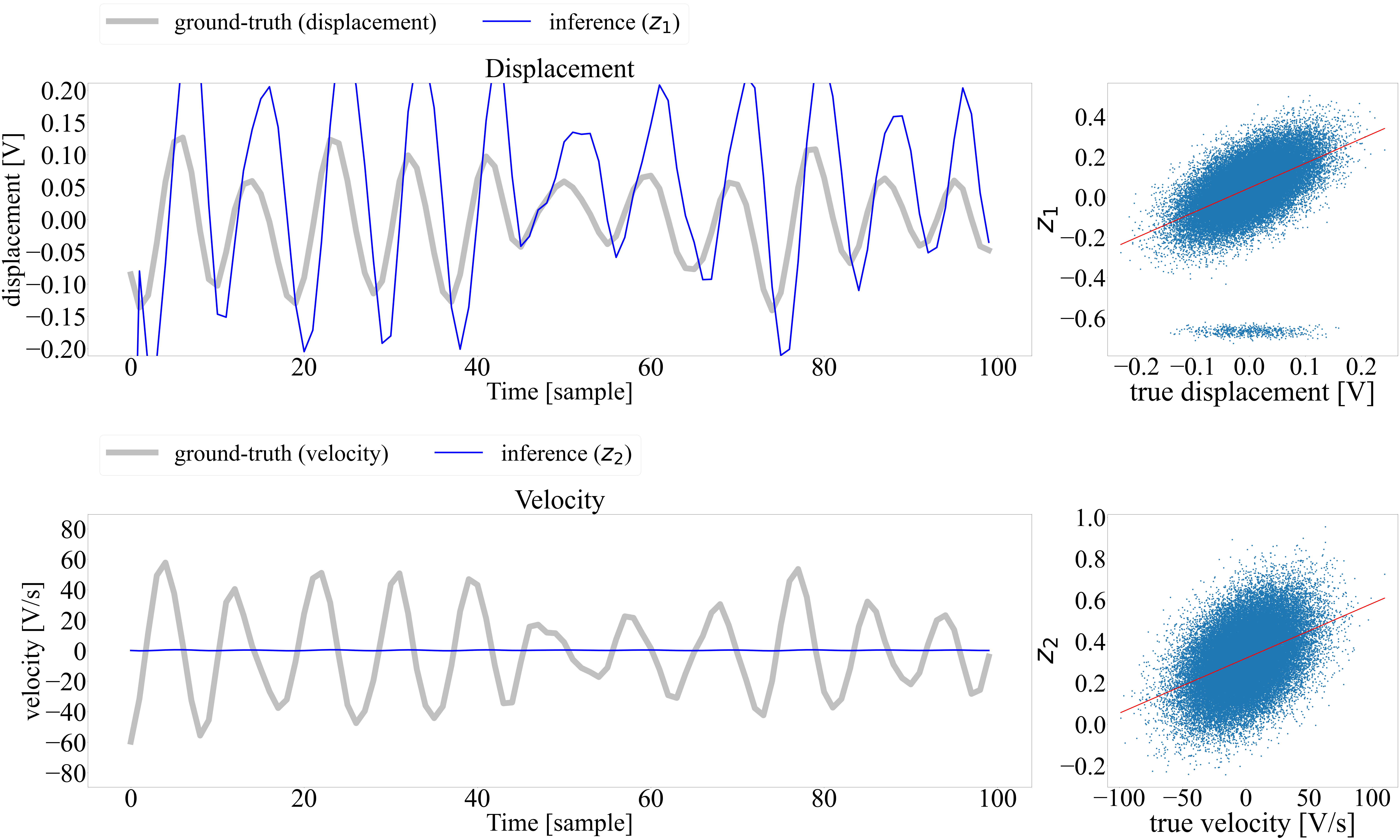}
  \caption{DMM}
  \label{fig:silverbox_expanded_dmm}
\end{subfigure}
\caption{Training results for the Silverbox benchmark problem. {\color{black} The velocities are the first-order difference of ``displacement-like" electric signals; the corresponding units are V/s. Right: The scatter plots of the ground-truth versus inferred latent states from corresponding models. The fitted linear regression lines are plotted in red.}}
\label{fig:silverbox_train}
\end{figure}

\begin{figure}[htpb]
\centering
\begin{subfigure}{1\textwidth}
  \centering
  \includegraphics[height=0.56\linewidth]{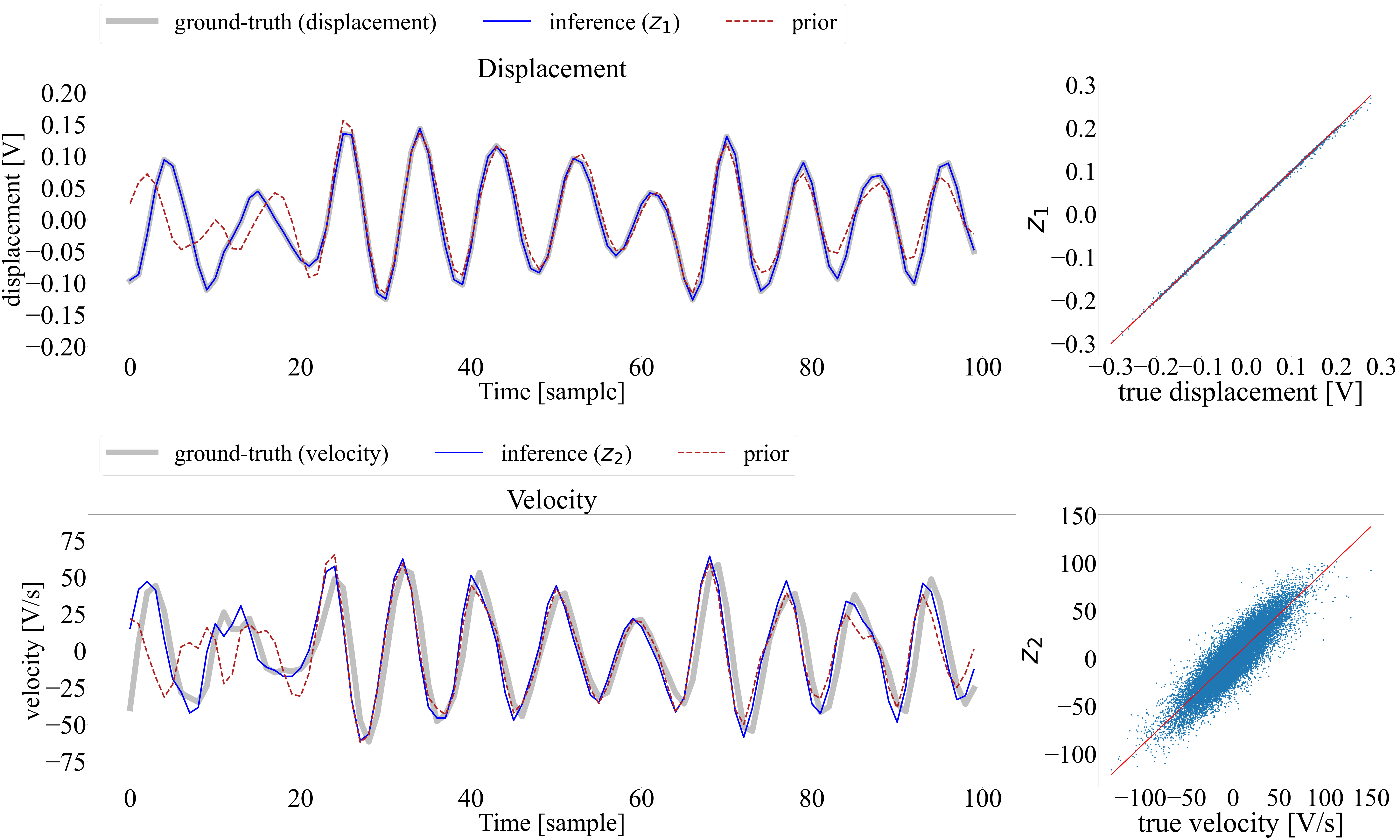}
  \caption{PgDMM}
  \label{fig:silverbox_expanded_test}
\end{subfigure}
\vfill
\begin{subfigure}{1\textwidth}
  \centering
  \includegraphics[height=0.56\linewidth]{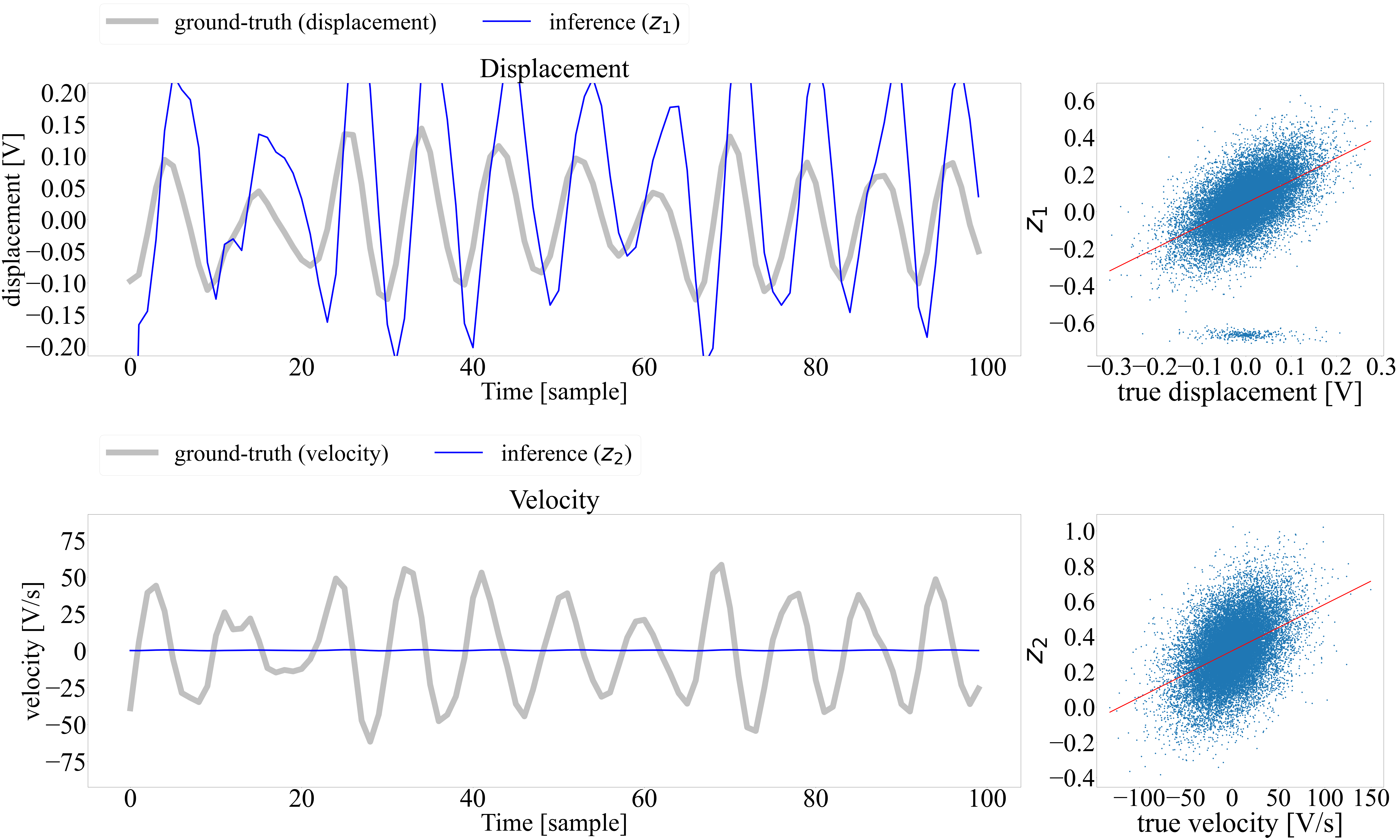}
  \caption{DMM}
  \label{fig:silverbox_expanded_dmm_test}
\end{subfigure}
\caption{Testing results for the Silverbox benchmark problem. {\color{black} Right: The scatter plots of the ground-truth versus inferred latent states from corresponding models. The fitted linear regression lines are plotted in red.}}
\label{fig:silverbox_test}
\end{figure}

\begin{figure}[t]
\centering
\begin{subfigure}{.5\textwidth}
  \centering
  \includegraphics[height=0.68\linewidth]{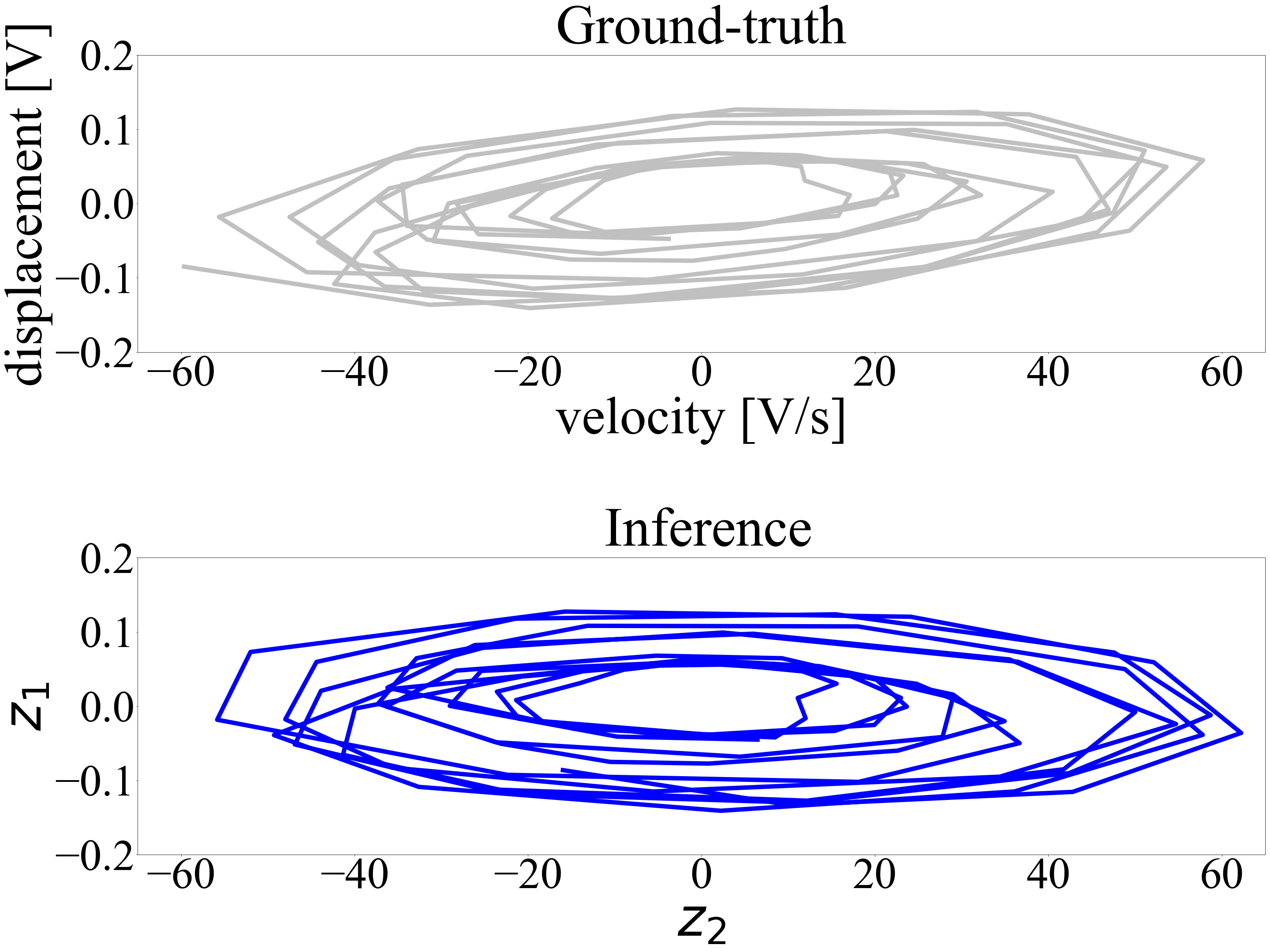}
  \caption{PgDMM}
  \label{fig:silverbox_phase}
\end{subfigure}%
\begin{subfigure}{.5\textwidth}
  \centering
  \includegraphics[height=0.68\linewidth]{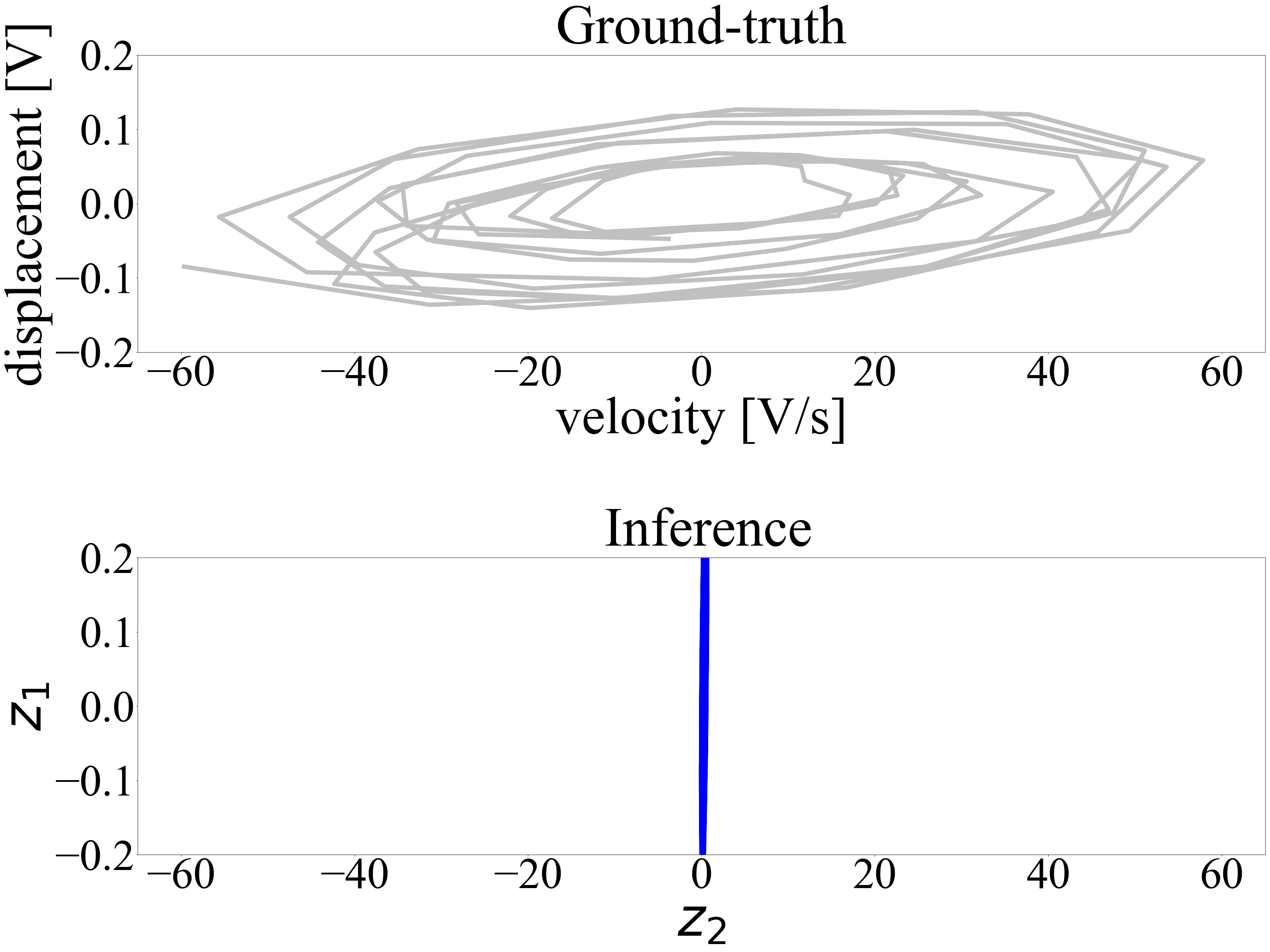}
  \caption{DMM}
  \label{fig:silverbox_phase_dmm}
\end{subfigure}
\caption{Phase portraits of training results. {\color{black} Top: Phase portraits of ground-truth displacements versus velocities. Bottom: Plots of inferred latent states $\textbf{z}_1$ versus $\textbf{z}_2$.}}
\label{fig:silverbox_train_phase}
\end{figure}

Figure \ref{fig:silverbox_train} and \ref{fig:silverbox_train_phase} offer a comparison, over a segment of the analysis window ($t = 37,201$ to $37,300$), between the learned latent variables ($z_1$ and $z_2$) and phase spaces of PgDMM and the conventional DMM scheme, both plotted against the ground-truth. From this comparison, we can deduce that the latent states inferred by our model fit the ground-truth very well, as the plots are almost overlapping. DMM, on the other hand, captures little information about the displacement (the phase seems correct) and delivers almost no information on the velocity state (the first-order derivative). {\color{black} As mentioned, the main drawback of the DMM, which constitutes a purely data-driven learning scheme, lies in that the latent space is imposed no physical structure. This implies that the latent states do not need to hold a physical interpretation, but instead represent some arbitrary features that the model finds beneficial for reconstruction and maximization of the observed data likelihood (i.e. the objective function ELBO). The limitation of the DMM in capturing velocity information has also been explored in existing literature \cite{karl2016deep}.} The lines labeled by ``prior'' shown in Figure \ref{fig:silverbox_train} denote the latent states generated by the physics-guided transition function $f_\textbf{phy}$ in PgDMM, as indicated in Eq.\eqref{eq:silver_phy} using the learned initial values.

\begin{figure}[t]
\begin{subfigure}{.5\textwidth}
  \centering
  \includegraphics[height=0.68\linewidth]{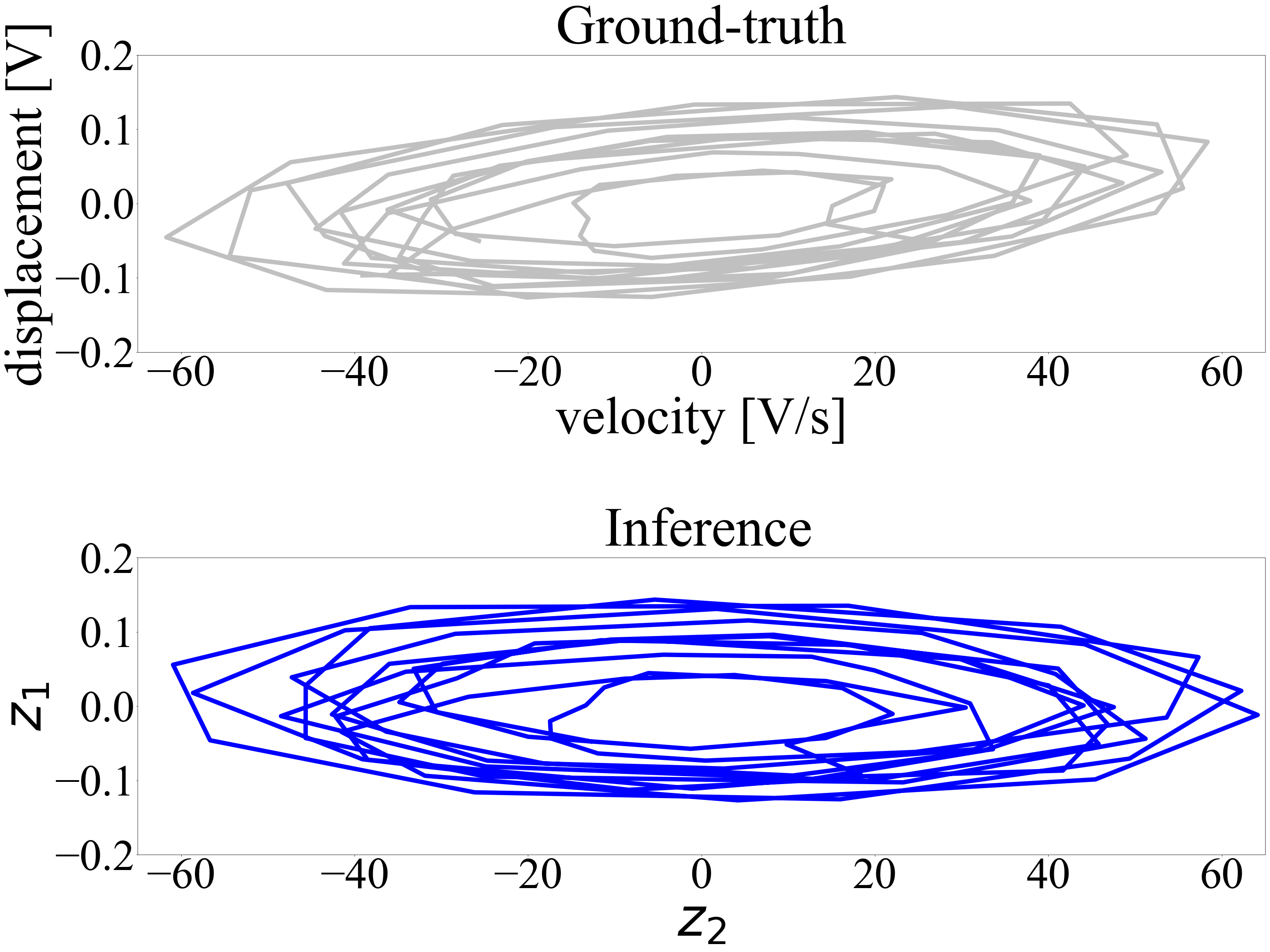}
  \caption{PgDMM}
  \label{fig:silverbox_phase_test}
\end{subfigure}%
\begin{subfigure}{.5\textwidth}
  \centering
  \includegraphics[height=0.68\linewidth]{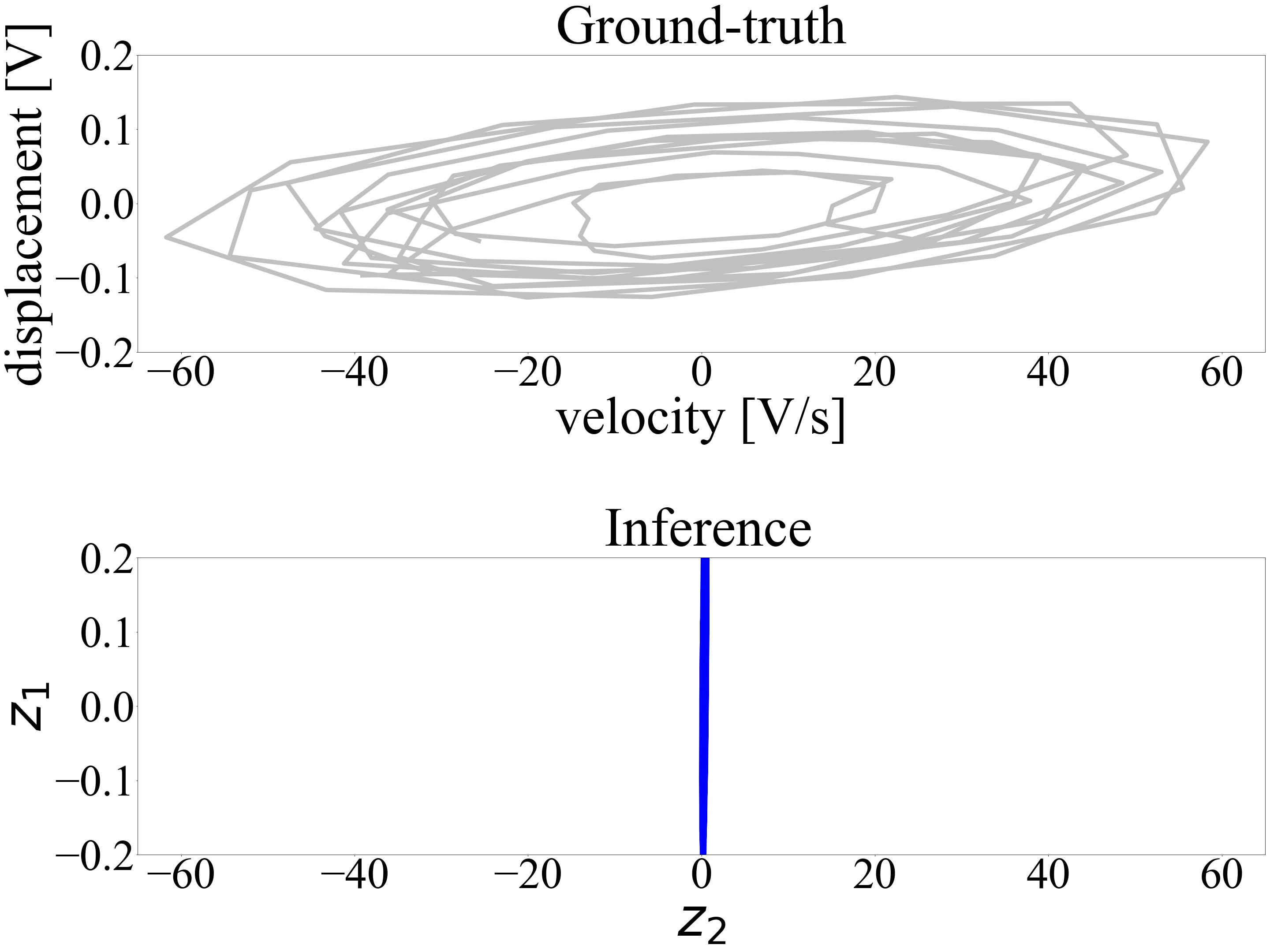}
  \caption{DMM}
  \label{fig:silverbox_phase_dmm_test}
\end{subfigure}
\caption{Phase portraits of testing results. {\color{black} Top: Phase portraits of ground-truth displacements versus velocities. Bottom: Plots of inferred latent states $\textbf{z}_1$ versus $\textbf{z}_2$.}}
\label{fig:silverbox_test_phase}
\end{figure}

The regression results in {\color{black} the right hand side of} Figure \ref{fig:silverbox_train} further validate that the latent states inferred by our model and the ground-truth exhibit a strong correlation, which indicates that the PgDMM is able to learn a physically interpretable latent space. The goodness-of-fit score, $R^2$, for the displacement is equal to $0.99994$ and for the velocity this result equals to $0.80059$, indicating a good fit to the ground-truth. Note that the measurements are only displacements, while a good score is also achieved in terms of the velocity estimate. On the other hand, the regression results for the DMM scheme yield $R^2$ values as low as $0.25324$ and $0.17760$ for displacement and velocity, respectively, which indicates a low correlation between the learned latent space and the ground-truth.

Similar results are obtained for the test dataset, as illustrated in Figures \ref{fig:silverbox_test} and \ref{fig:silverbox_test_phase}. Note that the scale of test data exceeds that of training data, thus the testing is extrapolative. The performance on the test dataset results is similar to the training dataset, indicating that a physics-guided model further favors extrapolation potential. We further compare the results using the root mean square error (RMSE) as a performance metric, which is computed as $\text{RMSE}=\sqrt{\frac{1}{T}\sum_{t=1}^T(\hat{\textbf{z}}_t-\textbf{z}_t)^2}$, where $\textbf{z}_t$ represents the ground-truth latent state and $\hat{\textbf{z}}_t$ represents the inferred latent state. The comparison of RMSE and $R^2$ is summarized in Table \ref{tab:results}, where the subscripts 1 and 2 represent the corresponding results for displacement and velocity respectively.

\begin{table}[htpb]\small
\caption{Summary of performance metrics for the Silverbox benchmark example} 
\centering 
\begin{tabular}{c ccccccccc} 
\hline\hline 
&\multicolumn{4}{c}{Training dataset} &\vline &\multicolumn{4}{c}{Test dataset}\\ [0.5ex]
Model &$R^2_1$ &$R^2_2$ &$\text{RMSE}_1$ &$\text{RMSE}_2$ &\vline &$R^2_1$ &$R^2_2$ &$\text{RMSE}_1$ &$\text{RMSE}_2$ \\ [0.5ex]
\hline 
DMM &0.25324 &0.17760 &0.12469 &22.9438  &\vline &0.24466 &0.17981 &0.12253 &22.3127 \\ [0.5ex] 
prior model &0.77326 &0.50881 &0.02680 &16.5039 &\vline &0.75448 &0.52281 &0.02687 &15.8457 \\ [0.5ex] 
PgDMM & 0.99994 & 0.80059 &0.00042 & 10.8810 &\vline &0.99984 &0.79552 &0.00067 &10.6107 \\ [0.5ex] 
\hline 
\end{tabular}
\label{tab:results}
\end{table}
\section{Conclusion}
We have introduced \textit{Physics-guided Deep Markov Models} (PgDMM) as a hybrid probabilistic framework for learning nonlinear dynamical systems from measured data. The proposed method combines a physics-guided model with a learning-based model, which aims at capturing the discrepancy between the physics-guided model and actual (monitored) dynamical system, to perform dynamical system identification. Via application of the proposed work on both synthetic and experimental data, we demonstrate that a physically disentangled representation can be obtained for the latent space, which is essential for extrapolative prediction capabilities of dynamical systems. The fusion reveals a considerable improvement when compared against the completely physics-based prior model and the completely learning-based DMM. {\color{black} While the described framework is not meant to tackle dynamical systems of multiple degrees of freedom, higher dimensional systems can be coupled with the framework described in \cite{tatsis2022hierarchical}, after a physics-based reduction is carried out.} Future work will consider further coupling of the physics-guided model with a nonlinear variant of Bayesian filters toward formulation of a hybrid inference model.

\section*{Acknowledgement}
The research was conducted at the Singapore-ETH Centre, which was established collaboratively between ETH Zurich and the National Research Foundation Singapore. This research is supported by the National Research Foundation, Prime Minister’s Office, Singapore under its Campus for Research Excellence and Technological Enterprise (CREATE) programme.


\bibliography{my_references}{}
\bibliographystyle{unsrt}

\newpage
\appendix
\section{Factorization of ELBO} \label{app:elbo}
With the assumed Markovian property for the interdependence of $\textbf{x}$ ($\textbf{x}$ is short for $\textbf{x}_{1:T}$) and $\textbf{z}$ ($\textbf{z}$ is short for $\textbf{z}_{1:T}$): $p(\textbf{z})=\prod_{t=1}^Tp(\textbf{z}_{t}|\textbf{z}_{t-1})$ and $p(\textbf{x}|\textbf{z})=\prod_{t=1}^Tp(\textbf{x}_{t}|\textbf{z}_{t})$, one can further decompose the ELBO function $\mathcal{L}(\theta,\phi;\textbf{x})$. We first factorize the approximate posterior $q_\phi(\textbf{z}|\textbf{x})$ with the mentioned Markovian behavior:
\begin{equation}
\begin{split}
q(\textbf{z}|\textbf{x})&=\prod_{t=1}^{T}q(\textbf{z}_t|\textbf{z}_{1:t-1},\textbf{x}),\\
& = \prod_{t = 1}^{T}q(\textbf{z}_t|\textbf{z}_{t-1},\textbf{x}),
\end{split}
\end{equation} 
where $q(\textbf{z}_1|\textbf{z}_0,\textbf{x})$ is simply $q(\textbf{z}_1|\textbf{x})$ as the initial condition inferred by measured data $\textbf{x}$. Then KL-divergence term can simplify as
\begin{equation}
\begin{split}
&\text{KL}(q(\textbf{z}_{1:T}|\textbf{x})||p(\textbf{z}_{1:T}))\\
&=\int_{\textbf{z}_1}\cdots\int_{\textbf{z}_T}q(\textbf{z}_{1:T}|\textbf{x}_{1:T})\log\frac{q(\textbf{z}_1)\cdots q(\textbf{z}_T|\textbf{z}_{T-1},\textbf{x}_{1:T})}{p(\textbf{z}_1)\cdots p(\textbf{z}_T|\textbf{z}_{T-1})}d\textbf{z}_T\cdots d\textbf{z}_1\\
&=\int_{\textbf{z}_1}\cdots\int_{\textbf{z}_T}q(\textbf{z}_{1:T}|\textbf{x}_{1:T})\sum_{t=1}^T\log\frac{q(\textbf{z}_t|\textbf{z}_{t-1},\textbf{x}_{1:T})}{p(\textbf{z}_t|\textbf{z}_{t-1})}d\textbf{z}_T\cdots d\textbf{z}_1\\
&=\sum_{t=1}^T\int_{\textbf{z}_1}\cdots\int_{\textbf{z}_T}q(\textbf{z}_{1:T}|\textbf{x}_{1:T})\log\frac{q(\textbf{z}_t|\textbf{z}_{t-1},\textbf{x}_{1:T})}{p(\textbf{z}_t|\textbf{z}_{t-1})}d\textbf{z}_T\cdots d\textbf{z}_1\\
&=\sum_{t=1}^T\int_{\textbf{z}_{t-1}}\int_{\textbf{z}_t}q(\textbf{z}_{t-1},\textbf{z}_t|\textbf{x}_{1:T})\log\frac{q(\textbf{z}_t|\textbf{z}_{t-1},\textbf{x}_{1:T})}{p(\textbf{z}_t|\textbf{z}_{t-1})}d\textbf{z}_td\textbf{z}_{t-1}\\
&\text{(irrelevant latent variables, $\textbf{z}_t$ for $t\notin\{t-1,t\}$, are integrated out)}\\
&=\sum_{t=1}^T\int_{\textbf{z}_{t-1}}\int_{\textbf{z}_t}q(\textbf{z}_{t-1}|\textbf{x}_{1:T})q(\textbf{z}_t|\textbf{z}_{t-1},\textbf{x}_{1:T})\log\frac{q(\textbf{z}_t|\textbf{z}_{t-1},\textbf{x}_{1:T})}{p(\textbf{z}_t|\textbf{z}_{t-1})}d\textbf{z}_td\textbf{z}_{t-1}\\
&=\sum_{t=1}^T\int_{\textbf{z}_{t-1}}q(\textbf{z}_{t-1}|\textbf{x}_{1:T})(\int_{\textbf{z}_t}q(\textbf{z}_t|\textbf{z}_{t-1},\textbf{x}_{1:T})\log\frac{q(\textbf{z}_t|\textbf{z}_{t-1},\textbf{x}_{1:T})}{p(\textbf{z}_t|\textbf{z}_{t-1})}d\textbf{z}_t)d\textbf{z}_{t-1}\\
&=\sum_{t=1}^T\mathbb{E}_{q(\textbf{z}_{t-1}|\textbf{x}_{1:T})}[\text{KL}(q(\textbf{z}_t|\textbf{z}_{t-1},\textbf{x}_{1:T})||p(\textbf{z}_t|\textbf{z}_{t-1}))].
\end{split}
\end{equation}
Substitute the above result into ELBO $\mathcal{L}(\theta,\phi;\textbf{x})$, the objective function now decomposes as:
\begin{subequations}
	\begin{equation}
	\begin{split}
	\mathcal{L}(\theta,\phi;\textbf{x}) & = \mathbb{E}_{q}[\log p(\textbf{x}|\textbf{z})] -\text{KL}[q(\textbf{z}|\textbf{x})||p(\textbf{z})] \\
	& = \sum_{t=1}^{T}\mathbb{E}_q\log p(\textbf{x}_t|\textbf{z}_t) -\text{KL}[q(\textbf{z}|\textbf{x})||p(\textbf{z})]\\
	& = \sum_{t=1}^{T}\mathbb{E}_q\log p(\textbf{x}_t|\textbf{z}_t) - \sum_{t=1}^T\mathbb{E}_{q(\textbf{z}_{t-1}|\textbf{x}_{1:T})}[\text{KL}(q(\textbf{z}_t|\textbf{z}_{t-1},\textbf{x}_{1:T})||p(\textbf{z}_t|\textbf{z}_{t-1}))]\\
	& = \sum_{t=1}^{T}\mathbb{E}_q\log p(\textbf{x}_t|\textbf{z}_t) - \sum_{t=1}^{T} \mathbb{E}_q\log q(\textbf{z}_t|\textbf{z}_{t-1},\textbf{x}_{1:T}) + \sum_{t=1}^{T}\mathbb{E}_q\log p(\textbf{z}_t|\textbf{z}_{t-1}),
	\end{split}
	\end{equation}
or more specifically (adding the specific parameterization):
\begin{equation}
\mathcal{L}(\mathcal{D};(\theta,\phi)) =  \sum_{t=1}^{T}\mathbb{E}_{q_{\phi}}[\log \underbrace{p_{\theta}(\textbf{z}_t|\textbf{z}_{t-1})}_{\text{transition}} + \log \underbrace{p_{\theta}(\textbf{x}_t|\textbf{z}_t)}_{\text{emission}} - \log \underbrace{q_{\phi}(\textbf{z}_t|\textbf{z}_{t-1},\textbf{x}_{1:T})}_{\text{inference}} ].
\end{equation}
\end{subequations}
\section{ELBO for Physics-guided Deep Markov Model}\label{app:ELBO}
In the following derivation, $\textbf{z}$ is abbreviated for $\textbf{z}_{1:T}$, and $\textbf{x}$ is abbreviated for $\textbf{x}_{1:T}$. The region $\Omega$ is defined as $\Omega=\{(\textbf{z}^\textbf{phy},\textbf{z}^\text{NN})|\alpha\textbf{z}^\textbf{phy}+(1-\alpha)\textbf{z}^\text{NN}=\textbf{z}\}$.
\begin{align*}
\log p(\textbf{x})&=\log\int p_\theta(\textbf{z})p_\theta(\textbf{x}|\textbf{z})d\textbf{z}\\
&=\log\int(\iint_\Omega p_\theta(\textbf{z}^\textbf{phy})p_\theta(\textbf{z}^\text{NN})d\textbf{z}^\textbf{phy}\,d\textbf{z}^\text{NN})p_\theta(\textbf{x}|\textbf{z})d\textbf{z}\\
&=\log\iiint_\Omega\frac{q_\phi(\textbf{z}|\textbf{x})}{q_\phi(\textbf{z}|\textbf{x})}p_\theta(\textbf{z}^\textbf{phy})p_\theta(\textbf{z}^\text{NN})p_\theta(\textbf{x}|\textbf{z})d\textbf{z}^\textbf{phy}\,d\textbf{z}^\text{NN}\,d\textbf{z}\\
&=\log\iiint_\Omega q_\phi(\textbf{z}|\textbf{x})\frac{p_\theta(\textbf{z}^\textbf{phy})}{q_\phi(\textbf{z}^\textbf{phy}|\textbf{x})}\frac{p_\theta(\textbf{z}^\text{NN})}{q_\phi(\textbf{z}^\text{NN}|\textbf{x})}p_\theta(\textbf{x}|\textbf{z})d\textbf{z}^\textbf{phy}\,d\textbf{z}^\text{NN}\,d\textbf{z}\\
&\geq\iiint_\Omega q_\phi(\textbf{z}|\textbf{x})\log[p_\theta(\textbf{x}|\textbf{z})\frac{p_\theta(\textbf{z}^\textbf{phy})}{q_\phi(\textbf{z}^\textbf{phy}|\textbf{x})}\frac{p_\theta(\textbf{z}^\text{NN})}{q_\phi(\textbf{z}^\text{NN}|\textbf{x})}]d\textbf{z}^\textbf{phy}\,d\textbf{z}^\text{NN}\,d\textbf{z}\\
&=\int q_\phi(\textbf{z}|\textbf{x})\log p_\theta(\textbf{x}|\textbf{z})\,d\textbf{z} \\
&+\iint q_\phi(\textbf{z}^\textbf{phy}|\textbf{x})q_\phi(\textbf{z}^\text{NN}|\textbf{x})\log[\frac{p_\theta(\textbf{z}^\textbf{phy})}{q_\phi(\textbf{z}^\textbf{phy}|\textbf{x})}\frac{p_\theta(\textbf{z}^\text{NN})}{q_\phi(\textbf{z}^\text{NN}|\textbf{x})}]d\textbf{z}^\textbf{phy}\,d\textbf{z}^\text{NN}\\
&=\mathbb{E}_{q_\phi}[\log p_\theta(\textbf{x}|\textbf{z})]+\mathbb{E}_{q_\phi}[\log\frac{p_\theta(\textbf{z}^\textbf{phy})}{q_\phi(\textbf{z}^\textbf{phy}|\textbf{x})}]+\mathbb{E}_{q_\phi}[\log \frac{p_\theta(\textbf{z}^\text{NN})}{q_\phi(\textbf{z}^\text{NN}|\textbf{x})}]\\
&=\mathbb{E}_{q_\phi}[\log p_\theta(\textbf{x}|\textbf{z})]-\big\{\text{KL}(q_\phi(\textbf{z}^\textbf{phy}|\textbf{x})||p_\theta(\textbf{z}^\textbf{phy}))+\text{KL}(q_\phi(\textbf{z}^\text{NN}|\textbf{x})||p_\theta(\textbf{z}^\text{NN}))\big\} \\
&=:\mathcal{L}_\textbf{phy}(\theta,\phi;\textbf{x}).
\end{align*}
Similar to the derivation of factorization for regular ELBO shown in Appendix \ref{app:elbo}, the ELBO for our model is further factorized as
\begin{align*}
\mathcal{L}_\textbf{phy}(\theta,\phi;\textbf{x})&=\sum_{t=1}^{T}\mathbb{E}_{q_\phi}\log p(\textbf{x}_t|\textbf{z}_t)\\
&-\big\{\sum_{t=1}^T\mathbb{E}_{q_\phi^\textbf{phy}(\textbf{z}_{t-1}^\textbf{phy}|\textbf{x}_{1:T})}[\text{KL}(q_\phi^\textbf{phy}(\textbf{z}_t^\textbf{phy}|\textbf{z}_{t-1}^\textbf{phy},\textbf{x}_{1:T})||p_\theta(\textbf{z}_t^\textbf{phy}|\textbf{z}_{t-1}^\textbf{phy}))]\\
&+\sum_{t=1}^T\mathbb{E}_{q_\phi^\text{NN}(\textbf{z}_{t-1}^\text{NN}|\textbf{x}_{1:T})}[\text{KL}(q_\phi^\text{NN}(\textbf{z}_t^\text{NN}|\textbf{z}_{t-1}^\text{NN},\textbf{x}_{1:T})||p_\theta(\textbf{z}_t^\text{NN}|\textbf{z}_{t-1}^\text{NN}))]\big\}.
\end{align*}
\section{Implementation Details of Case Studies}\label{app:arc}
{\color{black} As shown in Table \ref{tab:hyperparameter}, we tested several settings of the hyperparameters with increasing depth and width of the employed neural networks and listed their resulting ELBO values after training for 1,000 epochs. The table lists the results for the Silverbox benchmark case as an instance. The settings are listed in the format ``number of hidden layers $\times$ number of neurons in each layer'' for each neural network involved in the different models. It is noted that the performance is approximately at the same level for different settings and therefore we adopted commonly used medium depth and width according to the corresponding data dimension.

For the fairness of comparison between the proposed PgDMM and the original DMM scheme, the depth and width as well as the activation functions of the networks are kept the same for both. The difference lies in the introduction of the physics-based component and the physical prior models used for each case study are already detailed in respective subsections. For transparency and reproducibility purposes, we here further provide the details of the employed architectures, as follows. The architecture of each network is presented in the following format: hidden units of the first hidden layer + hidden units of the second hidden layer + number of outputs for mean value and number of outputs for covariance. Each number of units is followed by the activation function employed for that layer.
\begin{table}[H]\small
\caption{Hyperparameter tuning for the Silverbox benchmark example} 
\centering 
\begin{tabular}{ccccc ccc} 
\hline\hline 
&\multicolumn{3}{c}{Settings} &\vline &\multicolumn{1}{c}{PgDMM} &\vline &\multicolumn{1}{c}{DMM}\\ [0.5ex]
\# &Transition &Emission &GRU &\vline &ELBO &\vline &ELBO \\ [0.5ex]
\hline 
1 &$2\times20$ &$2\times20$ &$1\times50$ &\vline &1.3091 &\vline &1.1279 \\ [0.5ex] 
2 &$2\times50$ &$2\times50$ &$1\times100$ &\vline &1.3211 &\vline &1.1899 \\ [0.5ex] 
3 &$3\times50$ &$3\times50$ &$1\times100$ &\vline &1.2890 &\vline &1.1775 \\ [0.5ex] 
4 &$3\times100$ &$3\times100$ &$1\times200$ &\vline &0.8209 &\vline &1.1515 \\ [0.5ex] 
\hline 
\end{tabular}
\label{tab:hyperparameter}
\end{table}
\subsection{Dynamic Pendulum}
\begin{itemize}
    \item Input: 51 timesteps of $16\times16$ grayscale pixels
    \item Latent Space: 2 dimensions
    \item Inference Network ($\mathcal{NN}_\phi^i$) for $q^i_{\phi}(\textbf{z}^i_{t}|\textbf{z}^i_{t-1},\textbf{x}_{1:T}) = 
\mathcal{N}(\mathbf{\mu}_{\phi}^i(\textbf{z}^i_{t-1},\textbf{x}_{1:T}),\mathbf{\Sigma}_{\phi}^i(\textbf{z}^i_{t-1},\textbf{x}_{1:T}))$: GRU (for modeling dependence on $\textbf{x}_{1:T}$) and 128 tanh + 128 ReLU + 2 identity output ($\mu_\phi^i$) and 2 Softmax output ($\Sigma_\phi^i$)
    \item Transition Network ($\mathcal{NN}_1$) for $p(\textbf{z}_{t}^\text{NN}|\textbf{z}_{t-1}^\text{NN})\sim \mathcal{N}(\mathbf{\mu}_\text{NN}(\textbf{z}_{t-1}^\text{NN}),\mathbf{\Sigma}_\text{NN}(\textbf{z}_{t-1}^\text{NN}))$: 50 ReLU + 50 ReLU + 2 identity output ($\mu_\text{NN}$) and 2 Softmax output ($\Sigma_\text{NN}$)
    \item Emission Network for $p(\textbf{x}_{t}|\textbf{z}_{t})\sim \text{Bernoulli}(\mathcal{NN}_2(\textbf{z}_{t}))$: 128 ReLU + 128 ReLU + 256 sigmoid outputs
\end{itemize}
\subsection{Fatigue Crack Growth}
\begin{itemize}
    \item Input: 60 timesteps of 1 dimension
    \item Latent Space: 1 dimension
    \item Inference Network ($\mathcal{NN}_\phi^i$) for $q^i_{\phi}(\textbf{z}^i_{t}|\textbf{z}^i_{t-1},\textbf{x}_{1:T}) = 
\mathcal{N}(\mathbf{\mu}_{\phi}^i(\textbf{z}^i_{t-1},\textbf{x}_{1:T}),\mathbf{\Sigma}_{\phi}^i(\textbf{z}^i_{t-1},\textbf{x}_{1:T}))$: GRU (for modeling dependence on $\textbf{x}_{1:T}$) and 50 tanh + 50 ReLU + 1 identity output ($\mu_\phi^i$) and 1 Softmax output ($\Sigma_\phi^i$)
    \item Transition Network ($\mathcal{NN}_1$) for $p(\textbf{z}_{t}^\text{NN}|\textbf{z}_{t-1}^\text{NN})\sim \mathcal{N}(\mathbf{\mu}_\text{NN}(\textbf{z}_{t-1}^\text{NN}),\mathbf{\Sigma}_\text{NN}(\textbf{z}_{t-1}^\text{NN}))$: 20 ReLU + 20 ReLU + 1 identity output ($\mu_\text{NN}$) and 1 Softmax output ($\Sigma_\text{NN}$)
    \item Emission Network ($\mathcal{NN}_2$) for $p(\textbf{x}_{t}|\textbf{z}_{t})\sim \mathcal{N}(\mathbf{\mu}(\textbf{z}_{t}),\mathbf{\Sigma}(\textbf{z}_{t}))$: 20 ReLU + 20 ReLU + 1 identity output ($\mu$) and 1 Softmax output ($\Sigma$)
\end{itemize}
\subsection{Silverbox System Benchmark}
\begin{itemize}
    \item Input: 100 timesteps of 1 dimension
    \item Latent Space: 2 dimensions
    \item Inference Network ($\mathcal{NN}_\phi^i$) for $q^i_{\phi}(\textbf{z}^i_{t}|\textbf{z}^i_{t-1},\textbf{x}_{1:T}) = 
\mathcal{N}(\mathbf{\mu}_{\phi}^i(\textbf{z}^i_{t-1},\textbf{x}_{1:T}),\mathbf{\Sigma}_{\phi}^i(\textbf{z}^i_{t-1},\textbf{x}_{1:T}))$: GRU (for modeling dependence on $\textbf{x}_{1:T}$) and 100 tanh + 100 ReLU + 2 identity output ($\mu_\phi^i$) and 2 Softmax output ($\Sigma_\phi^i$)
    \item Transition Network ($\mathcal{NN}_1$) for $p(\textbf{z}_{t}^\text{NN}|\textbf{z}_{t-1}^\text{NN})\sim \mathcal{N}(\mathbf{\mu}_\text{NN}(\textbf{z}_{t-1}^\text{NN}),\mathbf{\Sigma}_\text{NN}(\textbf{z}_{t-1}^\text{NN}))$: 50 ReLU + 50 ReLU + 2 identity output ($\mu_\text{NN}$) and 2 Softmax output ($\Sigma_\text{NN}$)
    \item Emission Network ($\mathcal{NN}_2$) for $p(\textbf{x}_{t}|\textbf{z}_{t})\sim \mathcal{N}(\mathbf{\mu}(\textbf{z}_{t}),\mathbf{\Sigma}(\textbf{z}_{t}))$: 50 ReLU + 50 ReLU + 1 identity output ($\mu$) and 1 Softmax output ($\Sigma$)
\end{itemize}
}

\end{document}